\documentclass[a4paper,12pt,oneside]{article}
\usepackage[a4paper,top=3cm,bottom=3cm,left=3cm,right=3cm,marginparwidth=1.75cm]{geometry}
\usepackage[utf8]{inputenc}
\usepackage{graphicx}
\usepackage[english]{babel}
\usepackage{hyperref}
\usepackage{amsmath}
\usepackage{amsfonts}
\usepackage{amssymb}
\usepackage{minted}

\usepackage[numbers, square]{natbib}

\usepackage{mathtools}
\usepackage{bm}
\usepackage[caption=false]{subfig}
\usepackage{multirow}
\usepackage{url}

\newcommand{\mT}{\mathcal{T}}
\newcommand{\mU}{\mathcal{U}}

\newcommand{\mM}{\mathcal{M}}
\newcommand{\mN}{\mathcal{N}}

\title{\vspace{-2.5cm}{\bf Autonomous Unmanned Aircraft Systems for Enhanced Search and Rescue of Drowning Swimmers: Image-Based Localization and Mission Simulation}}

\author{
Sascha Emanuel Zell$^{1}$\thanks{Corresponding author: Sascha.Zell@b-tu.de}\,,
Toni Schneidereit$^{2}$,
Armin Fügenschuh$^{1}$,\\ Michael Breuß$^{2}$\\[2ex]
\small $^{1}$ Chair of Engineering Mathematics and Numerical Optimization \\ 
\small Brandenburg University of Technology Cottbus--Senftenberg \\ 
\small Platz der Deutschen Einheit 1 \\ 
\small 03046 Cottbus, Germany \\[2ex]
\small $^{2}$ Chair of Applied Mathematics \\ 
\small Brandenburg University of Technology Cottbus--Senftenberg \\ 
\small Platz der Deutschen Einheit 1 \\ 
\small 03046 Cottbus, Germany
}

\date{}

\begin{document}

\maketitle

\begin{abstract}
    Drowning is an omnipresent risk associated with any activity on or in the water, and rescuing a drowning person is particularly challenging because of the time pressure, making a short response time important. Further complicating water rescue are unsupervised and extensive swimming areas, precise localization of the target, and the transport of rescue personnel. Technical innovations can provide a remedy: We propose an Unmanned Aircraft System (UAS), also known as a drone-in-a-box system, consisting of a fleet of Unmanned Aerial Vehicles (UAVs) allocated to purpose-built hangars near swimming areas. In an emergency, the UAS can be deployed in addition to Standard Rescue Operation (SRO) equipment to locate the distressed person early by performing a fully automated Search and Rescue (S\&R) operation and dropping a flotation device. In this paper, we address automatically locating distressed swimmers using the image-based object detection architecture You Only Look Once (YOLO). We present a dataset created for this application and outline the training process. We evaluate the performance of YOLO versions 3, 5, and 8 and architecture sizes (nano, extra-large) using Mean Average Precision (mAP) metrics mAP@.5 and mAP@.5:.95. Furthermore, we present two Discrete-Event Simulation (DES) approaches to simulate response times of SRO and UAS-based water rescue. This enables estimation of time savings relative to SRO when selecting the UAS configuration (type, number, and location of UAVs and hangars). Computational experiments for a test area in the Lusatian Lake District, Germany, show that UAS assistance shortens response time. Even a small UAS with two hangars, each containing one UAV, reduces response time by a factor of five compared to SRO.
\end{abstract}
%\begin{center}
\vspace{0.875cm}
%\end{center}

\section{Introduction}\label{sec:introduction}

Swimming is a popular recreational activity and, due to its positive effects on physical and mental health, a recommended form of physical rehabilitation \citep{petrescu2014effects}. Despite its advantages, swimming and water sports in general also carry risks, especially drowning. Contrary to the prevalent misconception that drowning involves convulsive, panicked movements, it is often a silent event in which victims try to conserve energy and keep their heads just above the surface of the water, which often prevents them from calling for help \citep{jalalifar2024enhancing}. In unsupervised swimming areas, survival often depends on quick intervention of bystanders (usually non-professionals), whose reactions range from inaction to alerting \textit{Emergency Medical Services (EMS)} to high-risk personal intervention \citep{Clemens2017Addressing}. While ocean beaches are typically supervised by professional lifeguards, the situation differs in inland waters such as lakes and rivers. In Germany, around 90\% of fatal drowning accidents occur in inland waters such as rivers, lakes, streams, ponds, and canals, rather than in coastal areas \citep{dlrg2023_stats}. Supervision by lifeguards in these areas is usually only provided by volunteers and only during peak times; many areas remain completely unguarded due to staff shortages. Consequently, there is a need for scalable, technologically innovative solutions to enhance swimmer safety in unsupervised inland waters requiring none to little additional personnel.\par%

% innovations categories
Existing swimmer safety enhancing inventions have been categorized into physical barriers, swimmer detection devices, signaling devices, flotation devices, equipment for special water sports, devices for preventing drowning in home sanitary wares, devices for scuba diving and other miscellaneous devices \cite{john2019design}. The \textit{Unmanned Aircraft System (UAS)} presented in this study is primarily a device for detecting swimmers and providing flotation, with potential synergies with other categories listed above. This research was conducted as part of the interdisciplinary research project \textit{RescueFly}, funded by the Federal Ministry of Transport, Germany, which involved developing and testing a prototype UAS for an inland water test area in the \textit{Lusatian Lake District} in Eastern Germany \cite{vonBeesten2024RescueFly}. The UAS consists of a fleet of \textit{Unmanned Aerial Vehicles (UAVs)}, colloquially known as \textit{drones}, which are stored in custom-built hangars near swimming areas in a \textit{drone-in-a-box} configuration. In an emergency, victims can alert a control center themselves using smart watches or \textit{Wearable Health-Monitoring Systems (WHMSs)}, or bystanders can do so via mobile phones or on-site emergency telephones installed as part of the project. In addition to the resources for \textit{Standard Rescue Operations (SROs)}, the control center can dispatch the UAS to conduct an automated \textit{Search and Rescue (S\&R)} operation. Equipped with onboard sensors and self-inflating flotation devices, the UAVs can provide rapid support and bridge the therapy-free interval while lifeboats approach.\par%

\begin{figure}[ht]
    \centering
    \subfloat[Water rescue UAV prototype \textit{BUDDY I}, equipped with a \textit{Restube} \cite{restube}, positioned on the take-off area of the UAV hangar. \textit{Photo: Björn-Steiger-Stiftung.}]{%
        \includegraphics[width=0.48\textwidth]{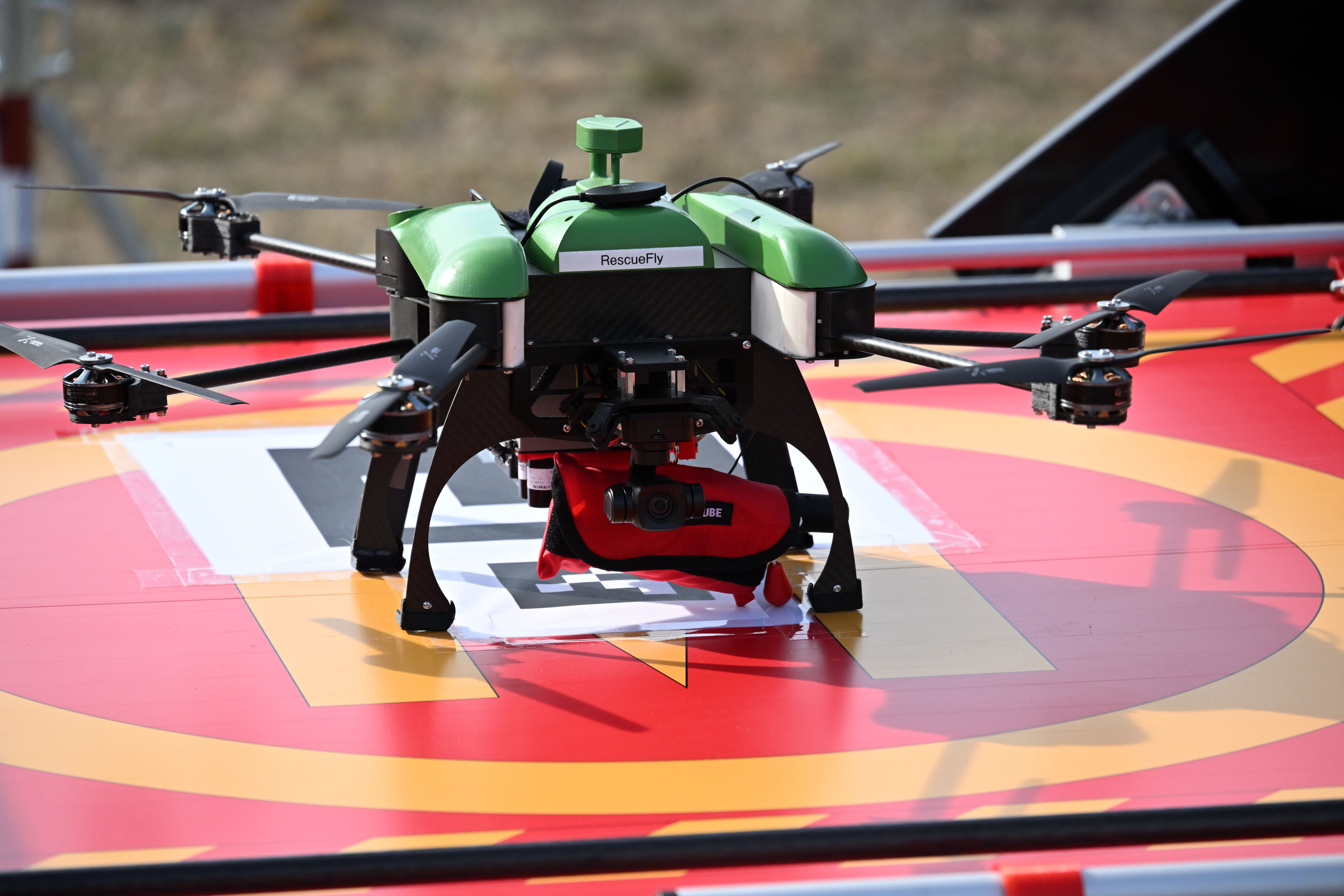}
        \label{fig:uav}
    }
    \hfill
    \subfloat[The RescueFly UAV hangar prototype, shown with open doors, equipped with a single UAV. \textit{Photo: Ralf Schuster, BTU Cottbus-Senftenberg.}]{%
        \includegraphics[width=0.48\textwidth]{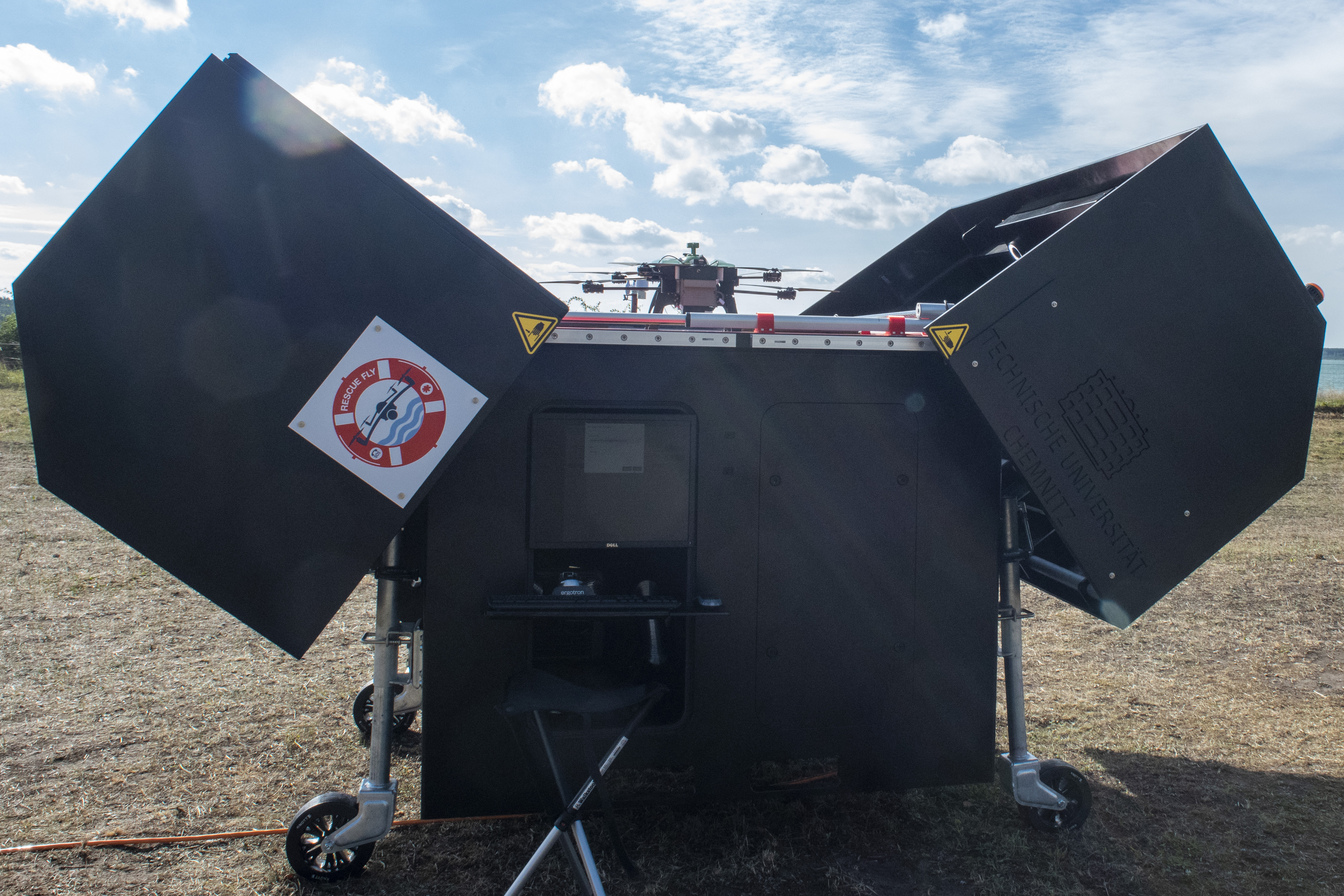}
        \label{fig:hangar}
    }
    \caption{A UAS consisting of water rescue UAV (left) and hangar (right).}
    \label{fig:rescuefly}
\end{figure}

Figure~\ref{fig:uav} shows \textit{BUDDY I}, a prototype water rescue UAV  manufactured by MINTMASTERS GmbH and equipped with an inflatable flotation device \textit{Restube} \cite{restube}, which was developed and tested as part of the \textit{RescueFly} project. The project also deployed the UAV hangar \textit{dRack}, shown in Figure~\ref{fig:hangar}, manufactured by DELTA-Fluid Industrietechnik GmbH and additionally equipped with optical sensors, computing hardware, and algorithms for automated post-flight checks \cite{heller2024}.

% add rescuefly pictures (maybe some new ones)

% describe paper focus (YOLO + simulation)
This paper is the second part of a two-part series. Part 1 introduced a \textit{Mixed-Integer Linear Program (MILP)} for in-advance offline S\&R flight planning of UAV swarms and another MILP for spatial UAS location-allocation optimization. In this second part, we extend that work by using the MILP results to run a \textit{Discrete-Event Simulation (DES)} of UAS-based S\&R operations. Another DES model enables the estimation of average SRO response times. Together, this forms an evaluation framework for the performance of different UAS configurations, quantifying the impact of additional resources on the response time. Furthermore, we present a real-time image-based swimmer localization approach, utilizing the \textit{You Only Look Once (YOLO)} neural network architecture. Due to limited available training data, we introduce a new dataset enriched with synthetic images and compare the performance of several YOLO versions and architecture sizes.

% describe paper structure
The remainder of this paper is organized as follows: Section~\ref{sec:lit-review} reviews relevant literature on image-based swimmer detection and simulation of water rescue. Section~\ref{sec:image-recognition} details the YOLO-based detection of distressed swimmers and Section~\ref{sec:sim} presents simulation frameworks for SRO- and UAV-based water rescue, respectively. Section~\ref{sec:case-study} comprises a case study conducted in the Lusatian Lake District in Germany. Finally, Section~\ref{sec:conclusions} concludes with a summary, potential extensions, and future research directions.

% ---------------------------------------------------------------
% Section: Literature Review
% ---------------------------------------------------------------

\section{Literature Review}\label{sec:lit-review}

This section reviews the most relevant literature on two main topics: image-based detection approaches for S\&R operations, and simulation-based approaches for UAV-based S\&R, particularly focusing on water rescue.

\subsection{Image-based Detection of Distressed Swimmers}

Image-based detection of humans in S\&R operations is impacted by several factors, including the availability and quality of training data, the categorization of human behaviors, and model robustness. Since most S\&R operations involve humans, the collection and use of sensitive training data raises ethical and legal questions, especially regarding data privacy. These considerations are particularly important in the \textit{European Union (EU)}, where data protection regulations are stringent \cite{schwartz2019global}. However, a variety of datasets have already been created to support human detection in S\&R context, such as \textit{VisDrone} \cite{9573394}, \textit{SeaDronesSee} \cite{Varga_2022_WACV}, \textit{UMA-SAR} \cite{MoralesUMASAR:2021}, and \textit{SARD} \cite{ahxm-k331-21}. Frequently used detection frameworks for these tasks include variants of \textit{YOLO} \cite{YOLOv1,YOLOv1-v8}, \textit{Single Shot Detectors (SSDs)} \cite{10.1007/978-3-319-46448-0_2} and \textit{Region-based Convolutional Neural Networks (R-CNNs)} \cite{6909475,9993862}. \citet{9618700} introduced a dataset on human behavior in swimming pool environments, combining above-water and underwater images to support drowning detection research. The dataset includes three behavioral classes, swim, drown, and idle, with each image depicting only one person. All of the individuals recorded are male, between 18 and 21 years old, and predominantly of Middle Eastern ethnicity. The dataset comprises 47 above-water and 44 underwater videos with an average length of 57 minutes at 30 \textit{frames per second (FPS)}. Finally, CNN-based methods were used for scene classification and human pose estimation. \citet{9981298} introduced a multi-modal dataset for \textit{Wilderness Search and Rescue (WSAR)}, consisting of 55,942 UAV-captured visual-thermal image pairs, covering various wilderness terrains, seasons, lighting conditions, and camera positions, with participants being visible on trails, off trails and on dirt roads. Ages range from 11 to 70 years, but with a bias towards the 18 to 25 age group. Baseline detection results with YOLOv5 indicate that thermal images generally yield higher accuracy, mainly because of the environment. For example, in forested areas, the thermal representation of humans shows a high contrast to the surroundings. Another UAV image dataset was introduced by \citet{9369386}, featuring injured and exhausted humans in non-urban outdoor environments, recording nine actors aged 7 to 55 in different positions, both natural and staged, extracting a total of 1,981 images. Then, synthetic variants were created to simulate different weather conditions. These datasets and a modified subset of \textit{VisDrone} (restricted to the \textit{person} class) were used to train and evaluate deep neural networks including \textit{Cascade R-CNN}, \textit{Faster R-CNN}, \textit{RetinaNet} and \textit{YOLOv4}, with \textit{YOLOv4} showing the best results. \citet{MARTINEZALPISTE2021114937} proposed a smartphone-compatible detection framework for UAV-based S\&R operations, implementing a \textit{YOLOv3-tiny} \cite{9074315} model optimized for real-time on-device processing of UAV footage. They also introduced a dataset comprised of 26,740 images from the Scottish wilderness, capturing seven background types, varying distances, camera positions, environmental and lighting conditions, and participant appearances. The system achieved a detection rate of approximately 7 FPS on the utilized smartphone device. \citet{SwimLocFromCamera} developed a swimmer tracking pipeline for competitive sports, integrating distinct features such as lane ropes and pool edges to evaluate the impact of camera position and perspective. \citet{lygouras2019unsupervised} presented an unsupervised human detection approach for autonomous UAV-based S\&R operations in an open water environment, potentially dropping flotation devices to assist distressed swimmers. They introduced a dataset from UAV perspective balancing images of swimmers and objects such as boats 1 to 1 to reduce false positives. Utilizing 9,000 images in total, 4,500 of these containing humans of different appearance, numbers, and perspectives with different lighting conditions, image sizes, and ratios, they trained and evaluated two CNN-based object detectors, \textit{YOLOv3-tiny} \cite{9074315} and \textit{SSD MobileNetV2} \cite{app8091678}. Their results indicate strong YOLO performance, though opportunities remain for further improvements. Continuing, \citet{SRdetectiontracking} combined detection and tracking into one framework, labeling 5,000 images for training a \textit{faster R-CNN} \citep{fastRCNN,masRCNN} model with \textit{ResNet-101 V1} \citep{ResNet}, while tracking was realized with a \textit{Kernelized Correlation Filter (KCF)} tracker \cite{KCFTracker} combined with \textit{Histogram of Gradients (HOG)} descriptors \cite{HOG}. While detection precision and accuracy were persuasive, processing speed remained a bottleneck, which is important for real-time rescue.

\subsection{Water Rescue Simulation}

Simulation is a popular tool in \textit{Emergency Medical Service (EMS)} planning. Classical methods such as \textit{Queuing Theory} have limitations, especially assuming a single call type or a homogeneous pool of agents, making them less suitable for modeling complex EMS systems \cite{vanbuuren2015}. In contrast, simulation approaches offer more flexibility and have been applied using different methodologies such as \textit{Agent-based Simulation (ABS)}, \textit{Discrete-Event Simulation (DES)}, \textit{System Dynamics}, and \textit{Monte-Carlo Simulation (MCS)}. Although the literature on UAV-based inland water rescue simulation is limited, there are relevant studies on water rescue in general, particularly maritime S\&R, as well as UAV-based S\&R operations in a broader context. \citet{cicek2022} conducted a real-world simulation of a river rescue by deploying dummies and compared response times of \textit{UAV-based Rescue Operations (UAV-RO)} and SRO, concluding that UAV-RO located the target significantly faster due to the UAVs' capabilities to scan large areas faster. Similarly, \citet{seguin2018unmanned} simulated UAV-based maritime S\&R operations, comparing flotation device delivery times with standard lifeguard procedures. \citet{schon2023model} applied ABS in a case study on maritime S\&R in Sweden, with coordinated deployment of aircraft, helicopters, and UAVs, to examine how the model representation and resolution affect the \textit{Measures of Effectiveness (MoE)} in \textit{Systems of Systems (SoS)} evaluations. \citet{Ashrafi2024} developed a scalable and flexible ABS framework for SRO in the Arctic that accounts for environmental conditions. They used MCS to assess the impact of seasonal changes and rescue resources on total times to rescue in the Norwegian Barents Sea. \citet{ho2022research} optimized rescue plans for multiple distressed targets at sea, also taking into account operating costs, by applying  the \textit{Floyd-Warshall Algorithm} and \textit{Grey Relational Analysis (GRA)}. \citet{lee2022simulation} utilized MCS as a decision support tool for resource allocation and search strategy selection (parallel sweep or expansion square) for maritime S\&R in Portugal, considering a drift model with uncertainty propagation to incorporate target movement caused by currents and wind in the probability density maps for the target location. They found that the expansion square pattern outperformed parallel sweep. In contrast to most simulation approaches which are typically DES- or ABS-based, \citet{yang2021} employed SD simulation to model coordinated water traffic emergency rescue  for maritime accidents such as ship collisions, sinkings, or explosions. \citet{waharte2010supporting} used a grid-based 3D computer simulation approach to compare target localization times of UAV-supported S\&R operation using different search heuristics in a simplified test environment. Similarly, \citet{Alotaibi_2019} evaluated S\&R search algorithms after a natural disaster by comparing the percentages of survivors rescued. 

\subsection{Problem Statement and Contribution}

In summary, most studies to date have focused on maritime S\&R with SRO resources such as boats or helicopters, while unmanned and autonomous vehicles such as UAVs received comparatively little attention. Furthermore, inland waters such as lakes, rivers, and ponds have been largely overlooked or only addressed superficially in the literature. However, in S\&R operations, specialized image-based detection models of distressed swimmers, trained on environment-specific datasets, can be critical yet also of great benefit in achieving reliable and accurate predictions \cite{10.1007/978-3-032-00071-2_18}.

To address these gaps, we introduce two main contributions. First, we present an image-based detection approach of distressed swimmers including a novel UAV-captured dataset in an Eastern German lake environment, collected in full compliance with data privacy regulations of the EU. The dataset includes aerial footage of humans engaged in different activities swimming, standing, diving, floating, or simulating an emergency situation, recorded by UAV from a top-down perspective. We train YOLO models for real-time detection, classifying swimmers as \textit{probably okay} or \textit{probably not okay}. Second, we introduce DES models to simulate both SRO water rescue and UAV-based water rescue to quantify the impact of the UAS on the response time, with the latter model allowing incorporation of any time-discrete search algorithm.

% ---------------------------------------------------------------
% Section: YOLO detection
% ---------------------------------------------------------------

\section{Image-based Detection of Distressed Swimmers}\label{sec:image-recognition}

Accurate and reliable localization of distressed swimmers is essential in autonomous UAV-based S\&R operations as it ensures precise drop-off of flotation devices and provision of accurate navigation data to the rescue personnel.\par%

UAVs can be equipped with a variety of sensors, many of which are useful for S\&R scenarios. Sensors are generally classified as \textit{active} or \textit{passive}. Active sensors emit their own signal and measure the reflected response. Contrary, passive sensors detect energy emitted or reflected from external sources. Popular passive sensors include \textit{visual}, \textit{thermal}, and \textit{infrared (IR)} cameras, as well as \textit{spectrometers}, whereas popular active sensors include \textit{radio detection and ranging (radar)}, \textit{light detection and ranging (LiDAR)}, \textit{ultrasonic sensors (sonar)} or \textit{active IR} sensors \citep{yasin2020unmanned}. Both sensor types have distinct benefits and drawbacks in the context of distressed swimmer detection. Visual cameras are lightweight, compact, energy-efficient and easy to mount. However, extracting usable information from visual data requires notable computational resources, the \textit{Field Of View (FOV)} is limited and performance heavily depends on weather and lighting conditions \cite{yasin2020unmanned}. For the latter, IR cameras can support detection, as demonstrated by \citet{kim2012autonomous} for detecting cars in all lighting conditions. Active sensors, in contrast, are less affected by weather, have fast response times, induce less computational effort and can accurately measure distance and angular position \cite{yasin2020unmanned}. However, they also have drawbacks, for instance LiDAR cannot detect transparent objects \citep{yasin2020unmanned}, radar's low output resolution impedes exact object reconstruction \citep{kwag2004obstacle}. Ultimately, sensor selection must be context-oriented, in this case with regard to water rescue.\par%

For our study, we chose a visual camera as the primary sensor since inland water swimming is typically carried out in good weather conditions, during daylight, and in relatively calm waters. Under good lighting conditions and water clarity, submerged objects are visible up to a depth of several meters from top-down UAV perspective. Moreover, the difficulty lays not in detecting a swimmer itself but rather in identifying distressed swimmers. Under these circumstances, visual imaging combined with image-processing techniques provide the most promising balance of accuracy, practicality, and cost-effectiveness. To enable real-time detection and classification, we adopt the YOLO framework \cite{YOLOv1-v8}, which has worked well in similar applications.%\par%

\subsection{YOLO Object Detection Frameworks}\label{sec:Models}

To address the detection and classification task, we employ CNN-based real-time object detection frameworks of different YOLO models. We concisely outline the concept of YOLO using \textit{YOLO version 1 (YOLOv1)} as an example, followed by an overview of the specific versions used in our work, YOLOv3, YOLOv5, and YOLOv8 \cite{YOLOv1-v8}. \par%

\begin{figure}[!ht]
    \centering
    \includegraphics[width=\textwidth]{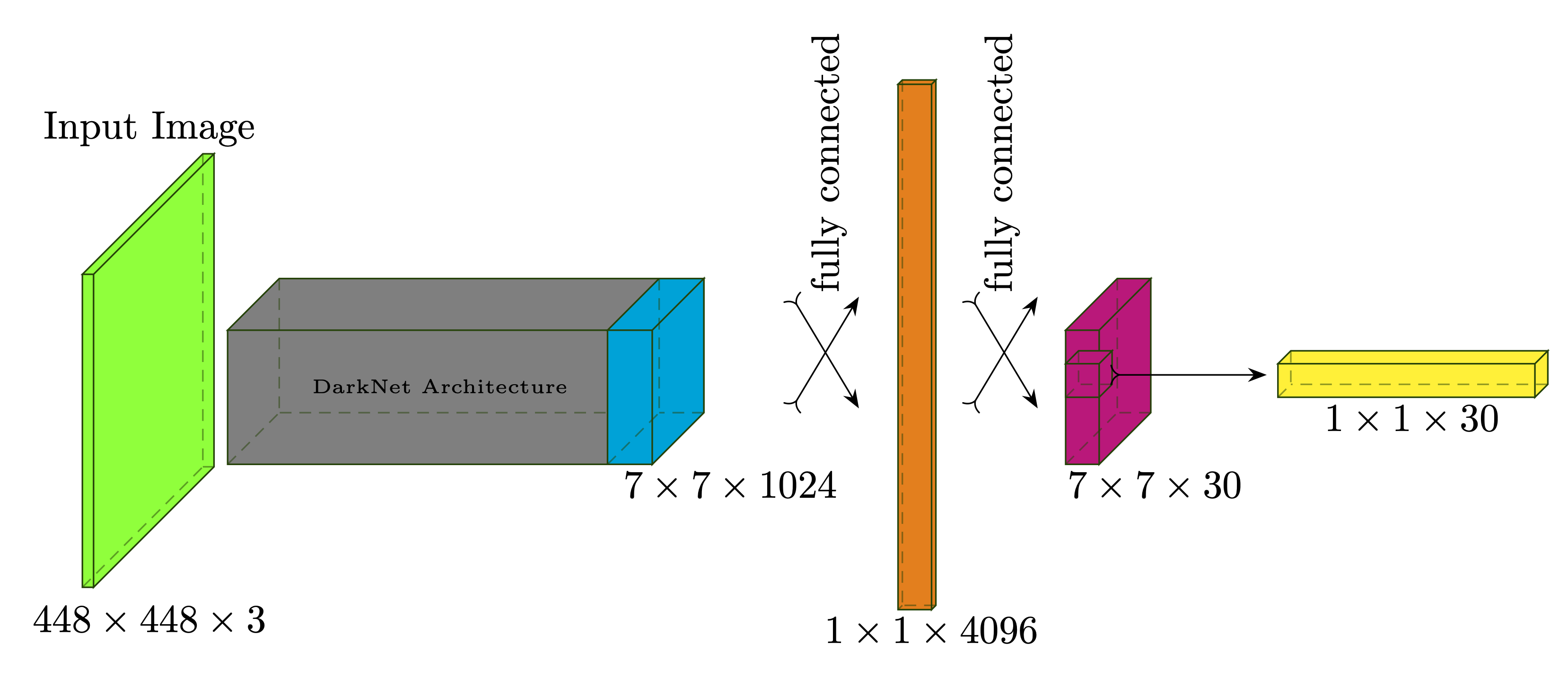}
    \caption{Schematic representation of the YOLOv1 architecture, showing the processing pipeline from input image (green) to detection output (pink/yellow) \cite{mohammadi2024Investigating}.}
    \label{fig:YOLOv1-arch}
\end{figure}

The YOLOv1 model, illustrated in Figure~\ref{fig:YOLOv1-arch}, is built upon the \textit{DarkNet} CNN architecture \citep{darknet1, darknet13} with $24$ convolutional layers, down-sampling an input image of size $448\times 448\times 3$ into a $7\times7\times1024$ feature tensor.
This effectively reduces the input image (illustrated in green in Figure~\ref{fig:YOLOv1-arch}) into a $7\times7$ grid (blue), with each cell encoding 1,024 features (gray). This flattened tensor is passed through a fully connected hidden layer (orange), producing a $7\times 7\times 30$ prediction tensor (pink). Each $1\times 1\times 30$ prediction vector (yellow) holds 20 conditional class probabilities based on the \textit{PASCAL Visual Object Classes (PASCAL-VOC)} dataset \citep{EEGWWZ2015} and two $1\times 1\times 5$ bounding box predictions consisting of box center coordinates $x$ and $y$, box height $h$ and width $w$, and \textit{confidence score} $\psi\in[0,1]$ \cite{YOLOv1}. YOLOv1 uses a compound loss function, combining \textit{localization loss}, \textit{confidence loss} and \textit{classification loss}, enabling real-time performance, as detailed by \citet{mohammadi2024Investigating}.\par%
YOLOv3~\cite{YOLOv3} significantly improves YOLOv1 regarding accuracy, speed, and detection robustness. It extends the CNN by incorporation of the Darknet-$53$ architecture \citep{darknet1, darknet13} with $53$ convolutional layers, integrating up-sampling and feature map interconnection via the \textit{Feature Pyramid Network (FPN) \citep{LDGHHB2016}}, which extracts high-level semantic information from deeper layers and fine-grained spatial details from earlier layers. The prediction tensor is extended to 80 conditional classes and three anchor bounding boxes on three different scales, resulting in a total of nine bounding boxes \cite{LMBBGHPRDZ2014}. These anchor boxes are derived during pre-processing via \textit{dimension clustering} and serve as prior shapes for the network to learn transformations to match ground-truth bounding boxes \cite{YOLOv2}.\par%

In 2020, \textit{Ultralytics LLC} introduced YOLOv5 \citep{YOLOv1-v8,YOLOv5,JocherYOLOv5byUltralytics2020}, a \textit{PyTorch}-based model with \textit{Cross Stage Partial Networks (CSPNet)} \citep{WLYWYH2019}, that improves accuracy and reduces model inference speed and model size by gradient change integration into feature maps. It retains FPN \citep{LDGHHB2016} and \textit{Path Aggregation Network (PANet)} \citep{LQQSJ2018} from YOLOv3, improving low-level feature propagation for localization precision.\par%

YOLOv8 \citep{YOLOv1-v8, Jocher_Ultralytics_YOLO_2023} introduces several architectural updates, altering convolution sizes and shifting towards \textit{Residual Networks (ResNet)} blocks \citep{he2015deep}. It adopts anchor-free detection, disposing the necessity to manually specify anchor boxes, resulting in better flexibility and computational efficiency.\par%

\subsection{Distressed Swimmer Detection Dataset}\label{sec:DataSe}

Collecting high-quality image data is a key requirement for training any object detection framework. In S\&R context, this typically involves humans, so video recordings become sensitive data under both ethical guidelines and strict data privacy regulations of the EU \cite{EUdataprivacy}. Consequently, the required amount and quality of data is not publicly accessible. Addressing this problem, we created our own dedicated dataset, as described below.

\subsubsection{Dataset Composition}\label{sec:datasetcomposition}

We collected a completely new dataset of human subjects exhibiting both standard swimming movements and partially staged distress behaviors in an inland water setting. The data acquisition took place at \textit{Lake Partwitz} in the \textit{Lusatian Lake District} in Eastern Germany over a two-day period, involving a total of 65 volunteers. All recordings were captured from UAV, in full compliance with the data privacy regulations of the EU and were also formally approved by the institution's ethics committee (see Declarations section for details).\par%

\begin{figure*}[!ht]
    \centering
    \includegraphics[width=\linewidth]{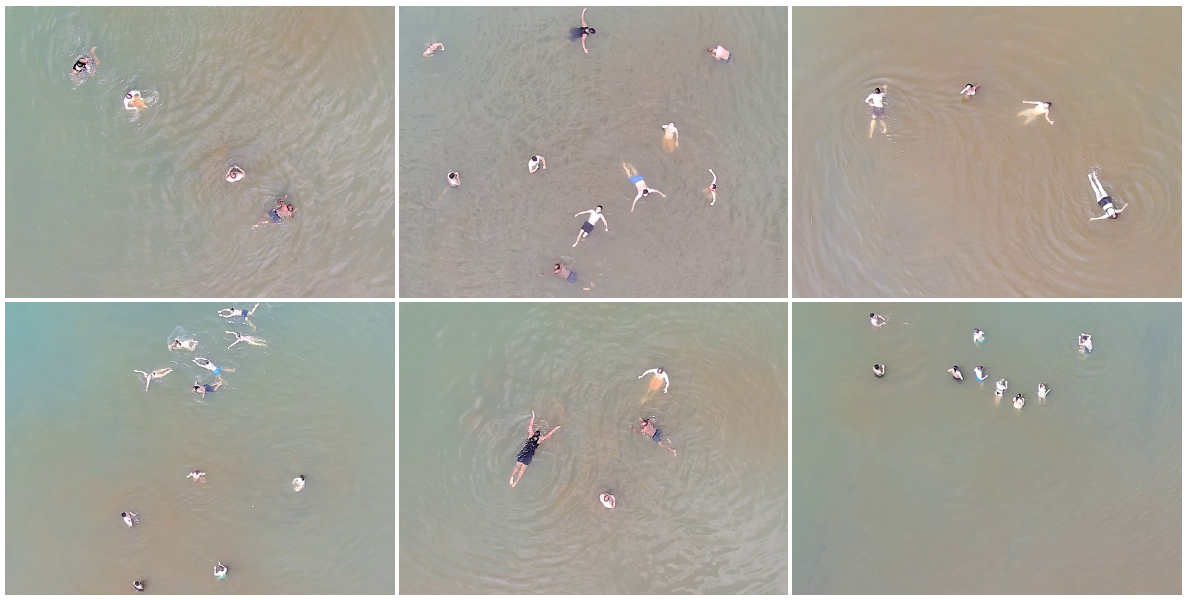}
    \caption{An overview of the UAV-captured dataset, in which heterogeneous groups of test subjects perform different standard swimming and distress behaviors.}
    \label{fig:droneScenes}
\end{figure*}

The resulting dataset consists of 65 GB of video recordings with a resolution of $2.560\times1.440$ pixels. Some example frames are shown in Figure~\ref{fig:droneScenes}, cropped from their original image size.
An important consideration in the dataset composition is the diversity of visible human participants. Scenes display groups of varying sizes, including male and female participants of several ethnic backgrounds. While this distribution represents the ethnic structure of Eastern Germany up to a reasonable degree, further expansion of the data set will be necessary, especially when the framework is deployed in other regions, to enhance fairness and equity, prevent discrimination, and to improve robustness and transferability of the resulting object detection models. For clarity, the term \textit{object} refers solely to detection targets within the algorithmic framework and does not imply objectification of human participants. Furthermore, the term \textit{swimmer} and \textit{person} will be used to refer to the individuals recorded.\par%

\begin{figure}[!ht]
    \centering
    \includegraphics[width=\linewidth]{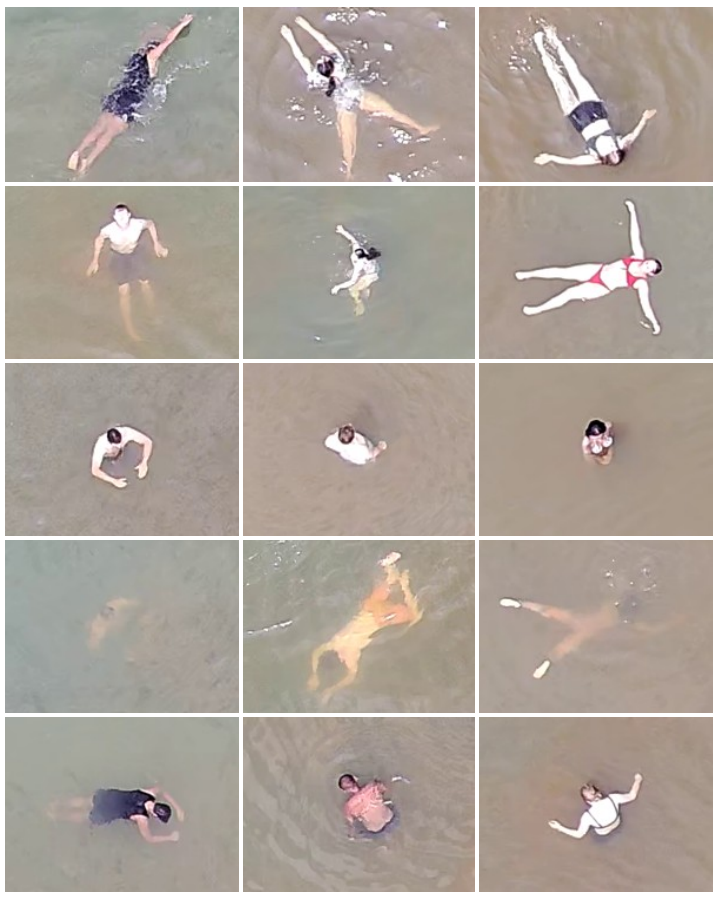}
    \caption{An overview of the different poses and movements in our dataset (from top row to bottom row); swimming, floating, standing, under water / diving, face-down floating.}
    \label{fig:SwimmStyles}
\end{figure}

As outlined in Section~\ref{sec:Models}, our approach is to train YOLO models, with the training data requiring a specific annotation format. For each image, a corresponding text file is generated holding all annotations or label names. This includes the object's class ID and the coordinates of the corresponding bounding box: the $x$ and $y$ coordinates of the center, along with its width $w$ and height $h$, all normalized relative to the image dimensions. We defined two target classes, intending to distinguish between individuals not appearing to be in distress and individuals potentially in distress. This resulted in the classes \textit{``prob. ok"} (the person is probably okay) and \textit{``prob. NOT ok"} (the person is probably not okay), respectively. This classification is based on visible behavior of humans in the recordings, which we broadly group into five categories, as shown in Figure~\ref{fig:SwimmStyles}. Row-wise, the top three rows refer to the categories of \textit{swimming}, \textit{floating}, and \textit{standing} and are assigned to the ``prob. ok." class. The two bottom rows refer to \textit{under water (diving)} and \textit{face-down floating} and are assigned to the "prob. NOT ok" class.

\subsubsection{Dataset Preparation}\label{sec:yolo_data_prep}

\begin{figure}[!ht]
    \centering
    \includegraphics[width=\linewidth]{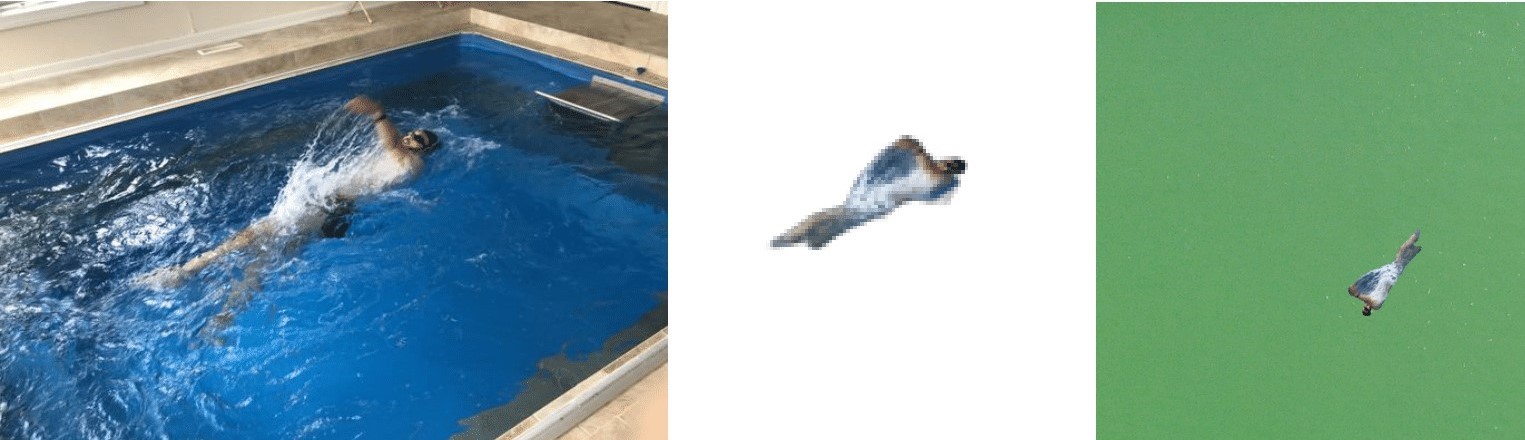}
    \caption{Synthetic image creation procedure: original swimmer image from Kaggle dataset (left), background-removed swimmer (middle), swimmer randomly placed on a lake background (right) \cite{mohammadi2024Investigating}.}
    \label{fig:SyntheticImage}
\end{figure}

From the UAV recordings, we ultimately derived a total of 18,518 labeled images. Achieving this volume was the result of an iterative cycle of labeling, training and correcting: Initially, we manually labeled 400 original images across the two classes. For training an object detection framework, this amount is insufficiently small. We approached this issue by synthetic image creation, as discussed by \citet{mohammadi2024Investigating}, and dataset augmentation using standard image processing techniques. To create synthetic images, we adopted a publicly available dataset from \textit{Kaggle} \cite{Kaggle}, featuring swimmers in an indoor pool environment performing various popular swimming styles. The indoor background was removed and the segmented swimmers were placed randomly onto the lake backgrounds from our UAV recordings. Figure~\ref{fig:SyntheticImage} illustrates this procedure.
As the location of the background-removed swimmer is known, bounding box labels were automatically extracted during image synthesis. However, a visual comparison with real-world UAV-captured imagery (see Figures~\ref{fig:droneScenes} and \ref{fig:SwimmStyles} versus Figure~\ref{fig:SyntheticImage}) reveals significant differences. Consequently, synthetic images were only employed during the labeling process by adding 2,000 synthetic images of the class ``prob. ok" to the 400 original images.\par%

\textit{Data augmentation} can improve the general representation of certain object features, so we employ some standard data augmentation techniques, manipulating the following image attributes \cite{YoloFactory}:
\begin{enumerate}%[label=(\arabic*)]
    \item \textit{Brightness:} Simulates varying lighting conditions
    \item \textit{Contrast:} Adjusts intensity differences
    \item \textit{Noise:} Mimics distance variations and sensor characteristics
    \item \textit{Motion blur:} Simulates camera movement
    \item \textit{Rotation:} Introduces variation in object orientation
    \item \textit{Translation:} Simulates different object positions
\end{enumerate}
Each transformation was applied once per image, expanding the 2,400 images by 14,400 augmented images, resulting in a total of 16,800 images. Then, a YOLOv5x model, the extra-large variant of VOLOv5, was trained on this dataset for 100 epochs with a batch size of 12. The trained model was then applied to 2,000 additional UAV-captured images that were not part of the initial dataset. On the results, we performed a manual label correction, by removing false detections (e.g. water reflections) and by correcting misclassifications between \textit{``prob. ok"} and \textit{``prob. NOT ok"}. This process was repeated with the now 2,400 labeled images and stopped at 18,518 labeled images for our final dataset. This procedure enables us to scale from a minimal unlabeled initial dataset to a large, high-quality labeled dataset, enriching real-world UAV-captured footage with synthetic and augmented data to enhance detection performance.

% ---------------------------------------------------------------
% Section: Simulation
% ---------------------------------------------------------------

\section{Simulation of UAV-based Water Rescue Operations}\label{sec:sim}

Simulation is a popular tool to validate models and assess the effectiveness of solutions for various EMS planning problems, such as facility location optimization, shift scheduling, resource allocation and dispatching strategies \cite{sorensen2010integrating, sudtachat2016nested, aboueljinane2013review}. In the following, two simulation models are presented. The first simulates water rescue SRO and the second simulates UAV-based water rescue of humans.

\subsection{Standard Rescue Operation Model Framework}\label{sec:sim-model_SRO}

In order to simulate the water rescue SRO in unsupervised swimming areas, the exact procedure must first be clarified. This is described by fire department personnel and control center staff in the respective area of the Lusatian Lake District in Germany as follows: Upon receiving an emergency call, the nearest available ambulance is dispatched and the responsible volunteer fire department is alerted, which then launches a rescue boat at a suitable water access point, typically a boat ramp. Together with the paramedics, the drowning victim is then approached by boat. We summarize this in the following four phases of the simulation:

\begin{enumerate}
    \item Alarm the nearest ambulance and fire brigade and send them to the  water access point closest to the accident site.
    \item Approach the selected water access point using the fastest path on the street network from the nearest ambulance and fire station, respectively.
    \item Prepare the lifeboat and optionally wait for paramedics to arrive at the location.
    \item Approach the accident location from the water access point using the fastest path on the water surface.
\end{enumerate}

To formally describe this process, we introduce the following notation: We define a set of fire stations $\mathcal{F}$ housing fire trucks, lifeboats and flotation devices and a set of rescue stations $\mathcal{R}$ housing ambulances. Further, we define a set of water access points $\mathcal{W}$ where lifeboats can be launched into the water. We suppose a static target position $\bm{\tau}\in\mathbb{R}^2$ over time. Now, we define travel times $\delta_{r,w}\in\mathbb{R}_{\geq 0}$ from $r\in\mathcal{R}$ to $w\in\mathcal{W}$ and travel times $\tilde{\delta}_{f,w}\in\mathbb{R}_{\geq 0}$ from $f\in\mathcal{F}$ to $w\in\mathcal{W}$ as well as travel times $\bar{\delta}_{w, \bm{\tau}}\in\mathbb{R}_{\geq 0}$ from $w\in\mathcal{W}$ to $\bm{\tau}$. We also define preparation times $\vartheta_F$, $\vartheta_R$ before starting the operation for the firefighters and paramedics, respectively. Finally, we determine the \textit{response time} or \textit{target approach time} $t^{\text{min}}_{\bm{\tau}}$. We identify the nearest water access point $\bar{w} = \text{arg}\min_{w\in\mathcal{W}} \bar{\delta}_{w, \bm{\tau}}$ to the target $\bm{\tau}$ as well as the nearest ambulance station $\bar{r} = \text{arg}\min_{r\in\mathcal{R}} \delta_{r,\bar{w}}$ and the nearest fire station $\bar{f} = \text{arg}\min_{f\in\mathcal{F}}\tilde{\delta}_{f,\bar{w}}$. Thus, we can compute

\begin{equation}\label{eq:sro_target_detection_time}
    t^{\text{min}}_{\bm{\tau}} = \max\{\vartheta_F+\tilde{\delta}_{\bar{f},\bar{w}}, \vartheta_R+\delta_{\bar{r},\bar{w}}\} + \bar{\delta}_{\bar{w}, \bm{\tau}}
\end{equation}

The first term $\max\{\vartheta_F+\tilde{\delta}_{\bar{f},\bar{w}}, \vartheta_R+\delta_{\bar{r},\bar{w}}\}$ is utilized to synchronize the forces since approaching the target on water from the water access point can only be started with both the firefighters and the paramedics on the lifeboat. We now employ a MCS approach by generating $N$ samples $\bm{\tau}_n$ of $\bm{\tau}$ and computing statistical parameters such as mean, median, and various quantiles of the estimated response time $t^{\text{min}}_{\bm{\tau}}$. Note that in the above simulation approach we made some simplifying assumptions especially assuming a stationary target with known location. In practice, the search operation will take additional time finding the exact location of the person. This indicates that the value of $t^{\text{min}}_{\bm{\tau}}$ can be interpreted as a kind of lower bound for the target approach time. This approach is further explored in the case study in section \ref{sec:SRO_sim_seenplatte}, which examines the specific choice of calculation methods for the proposed sets and parameters.

\subsection{UAV-based Water Rescue Model Framework}\label{sec:sim-model}

\begin{figure}[ht]
    \centering
    \includegraphics[width=0.3\textwidth]{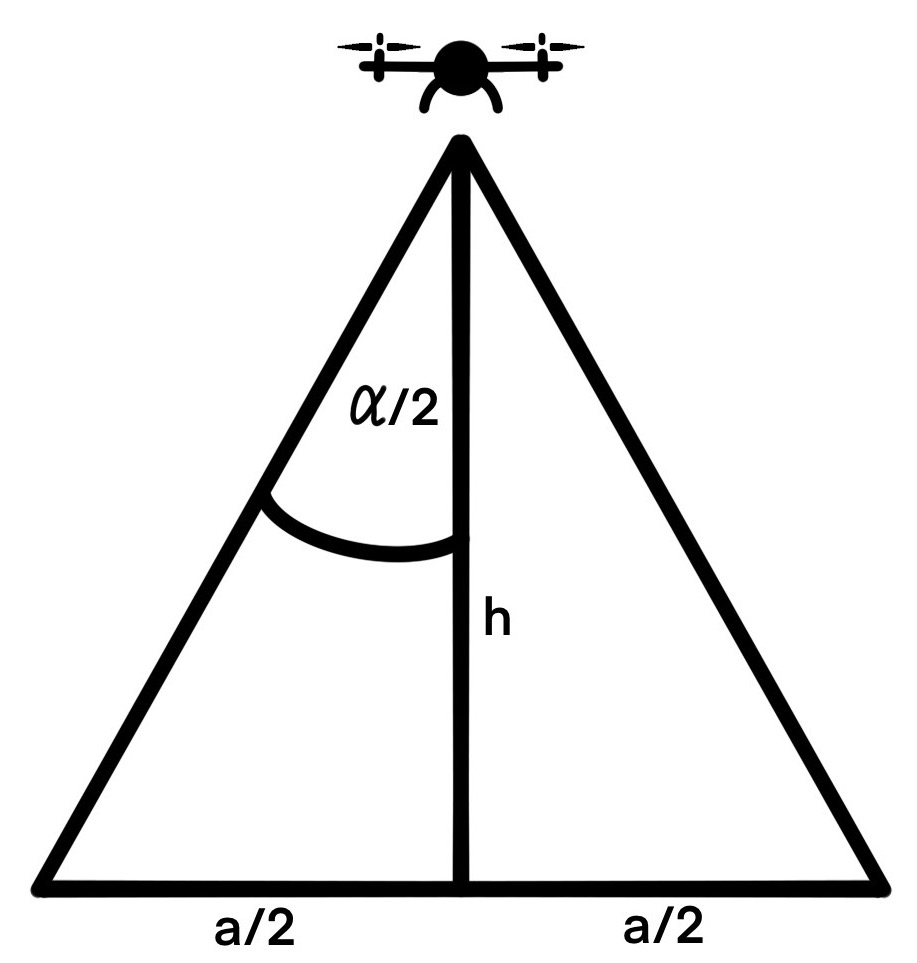}
    \caption{An illustration of the calculation of the horizontal FOV $a$ for a top-down UAV camera with angular FOV $\alpha$ and flight altitude $h$ (The lateral FOV is analogous).}\label{fig:drone_fov}
\end{figure}

We introduce a simulation framework that is designed to evaluate the performance of UAV-based search algorithms, which is particularly applicable to the proposed flight trajectory optimization model deployed by \citet{Zell2024} but also for every other search algorithm with discrete time steps. Here, time is discretized using a time horizon $T$, time step length $\Delta t_f\in\mathbb{R}_{>0}$, $n_f\in\mathbb{N}$ and the set $\mT_f = \{0,1,2,\ldots, n_fT \}$. We define the target position $\bm{\tau}_t = (\tau_{t}^x, \tau_{t}^y)^T\in\mathbb{R}^2$ and the UAV position $\bm{r}_{u,t} = (r_{u,t}^x, r_{u,t}^y)^T\in\mathbb{R}^2$ for UAV $u\in\mU$ at time step $t\in\mT_f$. The components of $\bm{r}_{u,t},\,\bm{\tau}_t$ refer to Cartesian $x$- and $y$-direction, respectively. Using the top-down camera's vertical and horizontal aperture angles $\alpha_u$, $\beta_u\in[0,\pi]$ and the UAV's flight altitude above ground $\eta_u\in\mathbb{R}_{>0}$, we can compute the vertical and horizontal FOVs $a_u,b_u\in\mathbb{R}_{>0}.$ Since the UAV's camera perspective is top-down, the flight altitude is sufficiently low to neglect the Earth's curvature, and the topographic profile of the landscape is negligible due to the flat water surface, we can use trigonometry  to obtain:
\begin{align}\label{eq:drone_fov}
    a_u/2 &= \eta_u \tan \alpha_u \\
    b_u/2 &= \eta_u \tan \beta_u 
\end{align}
This trigonometric relation is illustrated in Figure \ref{fig:drone_fov}. Note that a free-moving camera cannot be modeled this way and requires a more sophisticated approach. However, this would also mean a further degree of freedom in the optimization of the search mission, namely the consideration of the camera position in addition to the flight route planning.
We consider the air vectors $\bm{\iota}_{u,t} = (\iota_{u,t}^x, \iota_{u,t}^y)^T\in\mathbb{R}^2$ and ground vectors $\bm{\varrho}_{u,t} = (\varrho_{u,t}^x, \varrho_{u,t}^y)^T\in\mathbb{R}^2$ of UAV $u\in\mU$ at time step $t\in\mT_f$ as well as the wind vectors $\bm{w}_{t}\in\mathbb{R}^2$ with $\bm{\varrho}_{u,t} = \bm{\iota}_{u,t} + \bm{w}_{t}$.
We are now interested in computing the true heading angles $\xi_{u,t}$ to determine if the target is within the UAV's FOV. This is the angle between true north vector $\bm{q} = (0,1)^T$ and $\bm{\iota}_{u,t}$ which can be calculated as follows:
\begin{equation}\label{eq:true_heading_angle_computation}
    \cos \xi_{u,t} = \frac{\bm{q}\cdot \bm{\iota}_{u,t}}{\Vert \bm{q} \Vert\, \Vert \bm{\iota}_{u,t} \Vert} = \frac{\iota_{u,t}^y}{(\iota_{u,t}^x)^2 + (\iota_{u,t}^y)^2}.
\end{equation}
We utilize $\bm{\zeta} = (a/2,b/2)^T$ and two-dimensional rotation matrix 
\begin{equation}\label{eq:rot_mat}
    \bm{R}_{\gamma} = \begin{pmatrix}
        \cos \gamma & - \sin \gamma \\
        \sin \gamma & \cos \gamma
    \end{pmatrix}.
\end{equation}
To determine if a target is detected by UAV $u\in\mU$ at time step $t\in\mT_f$, we check if the following condition holds:
\begin{equation}\label{eq:target_detection_condition}
    - \bm{\zeta} \leq \bm{R}_{\xi_{u,t}} \cdot (\bm{r}_{u,t} -\bm{\tau}_t) \leq \bm{\zeta}.
\end{equation}
Rewriting Equation \eqref{eq:target_detection_condition}
component-wise, we obtain
\begin{align}\label{eq:target_detection_condition_componentwise}
    - \frac{a}{2} & \leq \cos\, \xi_{u,t}\, (r_{u,t}^x - \tau_{x,t}) - \sin\, \xi_{u,t}\, (r_{u,t}^y - \tau_{y,t}) \leq \frac{a}{2}, \\
    - \frac{b}{2} & \leq \sin \,\xi_{u,t}\, (r_{u,t}^x - \tau_{x,t}) + \cos \,\xi_{u,t}\, (r_{u,t}^y - \tau_{y,t}) \leq \frac{b}{2}.
\end{align}

Utilizing sets $V_{u, \bm{\tau}} \coloneqq\{t\in\mT_f \vert - \bm{\zeta} \leq \bm{R}_{\xi} \cdot (\bm{r}_{u,t} -\bm{\tau}_t) \leq \bm{\zeta} \}$ of time steps where the target is visible for UAV $u\in\mU$, we define the \textit{UAV's target approach time} for UAV $u\in\mU$ as

\begin{equation}
    t_{u, \bm{\tau}}^{\text{min}} \coloneqq \begin{cases}
        \min V_{u, \bm{\tau}}, & \text{ if } V_{\bm{\tau}} \neq \emptyset,  \\
        \infty, & \text{ else.}
    \end{cases}
\end{equation}
and the overall \textit{target approach time} or \textit{response time} as
\begin{equation}
    t_{\bm{\tau}}^{\text{min}} = \min_{u\in\mU} t_{u, \bm{\tau}}^{\text{min}}.
\end{equation}
The exact target location is typically unknown in advance, as knowing it would make the S\&R operation redundant. Since a drowning person is typically exhausted and almost immobilized, we assume a stationary target $\bm{\tau}=\bm{\tau}_t$ for all $t\in\mT_f$ and model it as a random variable. We may use different search methods $m\in\mM$ and want to compare them by their rate of successful to unsuccessful target detections $\chi_m\in[0,1]$ as well as their average finite detection time $\bar{t}_m\in\mathbb{R}_+$. We employ a MCS approach by generating $N$ samples $\bm{\tau}_n$ of $\bm{\tau}$ and computing target detection time $t_{\bm{\tau}_n,m}^{\text{min}}$ per method $m\in\mM$ for $n=1,\ldots, N$. This estimation method is meaningful only if we omit infinite target detection times, excluding cases where the target remains undetected. We define the sets $\mN^{det}_{m} = \{t_{\bm{\tau}_n,m}^{\text{min}}, n=1,\ldots N \vert t_{\bm{\tau}_n,m}^{\text{min}} < \infty \}$ and obtain
\begin{equation}\label{eq:expecation_average}
    \bar{t}_m = \frac{1}{\vert \mN^{det}_{m} \vert} \sum_{i\in\mN^{det}_{m}} i, %t_{n,m}^{\text{min}},
\end{equation}
\begin{equation}\label{eq:success_ratio}
    \chi_m = \frac{\vert \mN^{det}_{m} \vert}{N}.
\end{equation}
This approach allows us to quantify the gains in terms of response time and target detection rate when using additional UAVs and different search methods. Further, this allows us to compare different hangar location-allocation configurations.

% ---------------------------------------------------------------
% Section: Case Study
% ---------------------------------------------------------------

\section{Case Study and Scenario Analysis}\label{sec:case-study}

\begin{figure}[ht]
    \centering
    \subfloat[The many artificial lakes in the Lusatian Lake District, visualized with \textit{OpenStreetMap (OSM)}.]{%
        \includegraphics[width=0.48\textwidth]{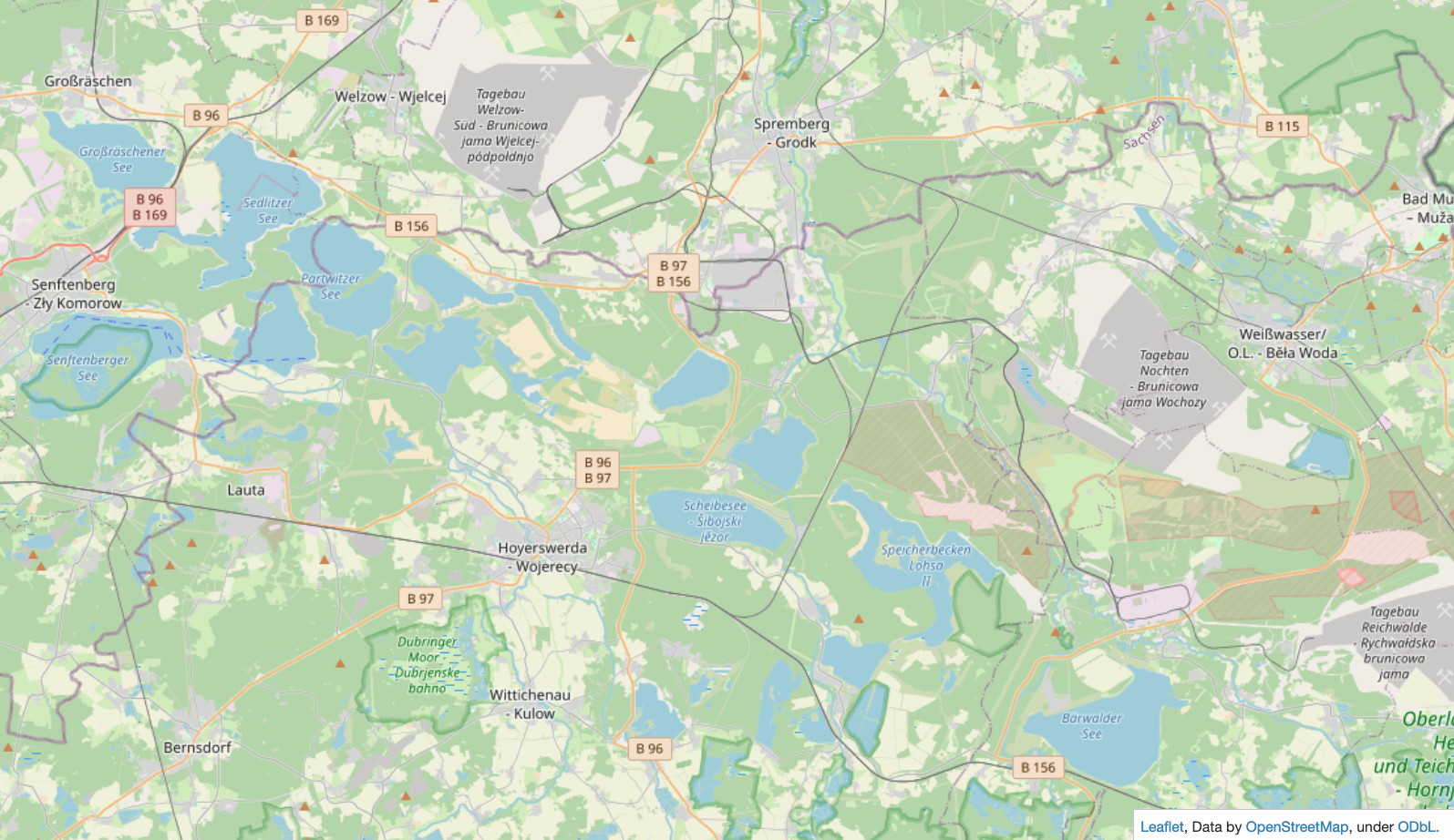}
        \label{fig:area_osm}
    }
    \hfill
    \subfloat[The active lignite open-cast mines (Welzow in the north, Nochten in the east) are clearly visible on ESRI's satellite map.]{%
        \includegraphics[width=0.48\textwidth]{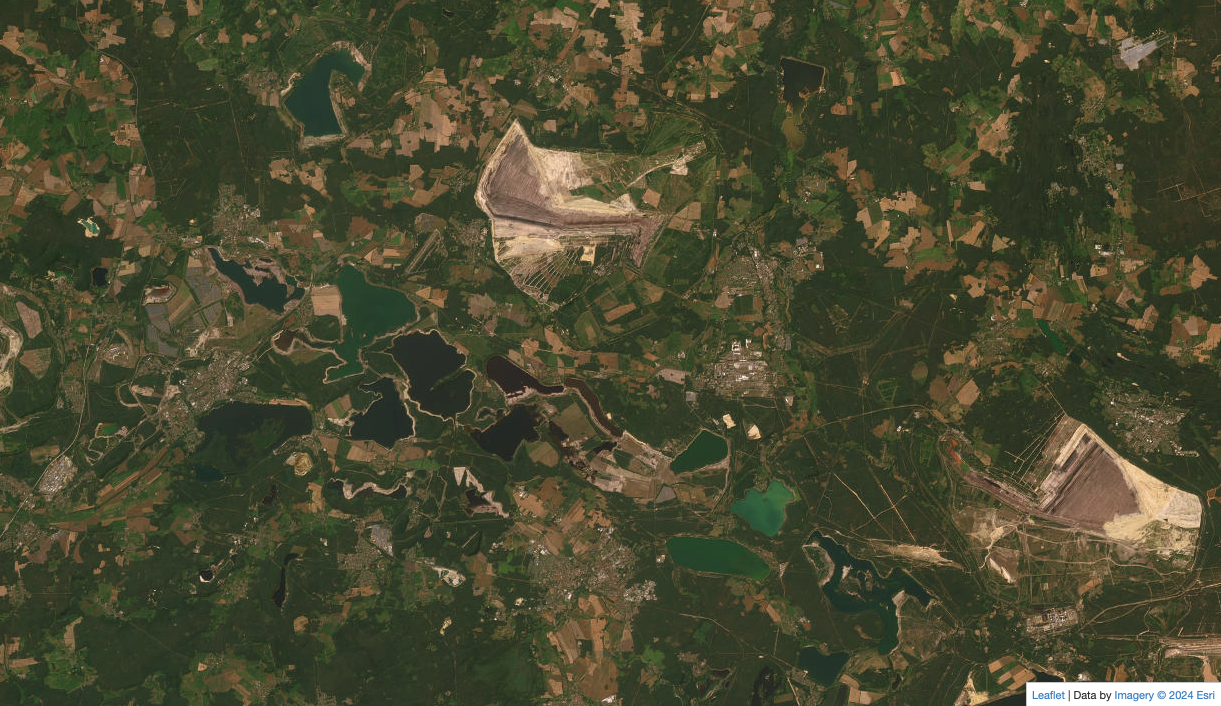}
        \label{fig:area_etsi}
    }
    \caption{Lusatian Lake District, Germany.}
    \label{fig:operational_area}
\end{figure}

The models presented earlier in Sections \ref{sec:image-recognition} and \ref{sec:sim} are applied to a test area in the \textit{Lusatian Lake District} in Eastern Germany. The entire \textit{Lusatian} region is an open-cast lignite mining area, where disused sites are flooded, creating Europe's largest artificial lake landscape. This is shown in Figure \ref{fig:operational_area}.
The aim of flooding these areas was to make them attractive to tourists, with swimming and other water sports proving very popular. However, large parts of these areas are not supervised by lifeguards. We therefore outline the benefits of UAS as additional support unit to enhance swimmer safety in this region.

% \subsection{YOLO Training and Evaluation}\label{sec:YoloTrainEval}
\subsection{Distressed Swimmer Detection and Localization}\label{sec:YoloTrainEval}

In order to train and evaluate the performance of an object detection framework, two different datasets are required. The training dataset contains training images and labels as well as validation images and labels. The latter typically comprises roundabout 10\% of the entire training dataset. For a reasonable performance testing, a separate evaluation dataset is needed, that contains images not seen by the framework during training. \par
We separated 1,018 images for the evaluation dataset and were left with 17,500 images for the training dataset. The latter has been augmented again, but in a slightly different way than before. In order to augment each image, we took the mentioned image manipulations from Section \ref{sec:yolo_data_prep} and randomly applied two of them at the same time to one image. In other words, for each original image we have one augmented image with two manipulated characteristics. In the end, the resulting 35,000 images were split to 31,500 training images and 3,500 validation images for the training dataset.% \par 

\subsubsection{Performance Evaluation}\label{sec:performance-eval}

To systematically assess the capabilities of the YOLO framework, we trained and compared three YOLO versions YOLOv3, YOLOv5, and YOLOv8, each with two different architecture sizes (nano and extra-large). Model performance was evaluated using the \textit{mean Average Precision (mAP)} metric, specifically the mAP@[.5] and mAP@[.5:.95] variants \cite{ObjDetMetrics}. Both metrics are based on the \textit{Intersection over Union (IoU)} $\phi\in[0,1]$, quantifying the overlap between the actual \textit{Ground Truth} bounding box and the predicted bounding box as a ratio. A prediction is considered a \textit{True Positive (TP)} if the IoU exceeds a predefined \textit{IoU threshold} $T^{\text{IoU}}$, else it is considered a \textit{False Positive (FP)}. The prediction is a \textit{False Negative (FN)} if a true object is missed. Table~\ref{tab:TPFPFN} summarizes those metrics for $T^{\text{IoU}}=0.5$.
\begin{table}[!ht]
    \centering
    \begin{tabular}{l|l|p{5.9cm}}
    Name & IoU value & Description  \\
    \hline
    True Positive (TP) & IoU $\ge 0.5$ & Prediction and ground truth are overlapping to a degree that is considered a correct detection. \\ 
    False Positive (FP) & IoU $< 0.5$ & Prediction and ground truth are  either in different locations or the overlap is not large enough to be considered a correct detection or there is a prediction without a given ground truth \\
    False Negative (FN) & IoU not defined & No Prediction for a given ground truth.
    \end{tabular}
    \caption{Comparison of possible cases for the prediction and the ground truth, based on the IoU.}
    \label{tab:TPFPFN}
\end{table}
It is important to note that the categories in Table \ref{tab:TPFPFN} are separately found with respect to each class. From those categories and with defining another \textit{detection threshold} $T^{\text{DET}}$ corresponding to YOLOs confidence score $\psi$, we derive additional metrics \textit{precision} $p = p(T^{\text{DET}}, T^{\text{IoU}})$ and \textit{recall} $r=r(T^{\text{DET}}, T^{\text{IoU}})$ as follows:
\begin{equation}
    p = \frac{\#TP}{\#TP + \#FP},
    \quad r = \frac{\#TP}{\#FP + \#FN}.
\end{equation}
Here, the annotation $\#(\ldots)$ denotes the respective number of obtained values in the evaluation dataset. By fixing IoU threshold $T^{\text{IoU}}$ and varying detection threshold $T^{\text{DET}}$, we obtain multiple \textit{precision-recall pairs} $\bigl(p(T^{\text{DET}}, T^{\text{IoU}}),$ $ r(T^{\text{DET}}, T^{\text{IoU}})\bigl)$ that are used to compute the \textit{Average Precision (AP)} $\bar{p}$, the area under the so-called \textit{precision-recall curve} $\hat{p}(r):[0,1] \rightarrow[0,1]$, which views precision $p$ as a function of recall $r$:
\begin{equation}
    \bar{p} = \int_{r=0}^1 \hat{p}(r)\,dr
\end{equation}
We use the commonly specified \textit{AP@[.5]}, using an indulgent IoU threshold $T^{\text{IoU}} = 0.5$ to primarily indicate if an object is detected at all. We will also employ \textit{AP@[.5:.95]}, where the AP is averaged over IoU thresholds $T^{\text{IoU}}_k = 0.5, 0.55, 0.6, \ldots, 0.95$, which is stricter, evaluating both detection and localization accuracy. When evaluating multiple classes, the individual APs are averaged over all classes to obtain the mAP. While \textit{mAP@[.5]} emphasizes object presence, \textit{mAP@[.5:.95]} provides a more comprehensive measure for localization accuracy.

\subsubsection{Experimental Setup}\label{sec:exp_setup}

The deployed computational devices and their corresponding hardware components are specified in Table~\ref{tab:DeviceSpecification} and are in the following referred to as \textit{``PC"} and \textit{``Laptop"}. Every YOLO model was trained with 200 epochs and a batch size of 12. We employed the nano (-n) and the extra lage (-x) architecture sizes of the YOLO model, with the nano versions being lightweight with focus on efficiency on limited hardware and the extra-large versions containing substantially more layers and parameters, trading speed for accuracy \citep{YOLOv1-v8}.
\begin{table}[!ht]
    \centering
    %\begin{tabular}{c|c|c}
    \begin{tabular}{lp{4.5cm}p{4.5cm}}
    \hline\noalign{\smallskip}
    Device & PC & Laptop  \\
    %\hline
    \noalign{\smallskip}\hline\noalign{\smallskip}
    CPU    & 32 × Intel Xeon Gold 6346 CPU & 16 × 13th Gen Intel Core i7-1360P\\[0.5ex]
    %\hline
    GPU    & RTX A2000 GPU 12GB, 3328 CUDA cores & RTX A500 Laptop GPU 4GB, 2048 CUDA cores \\
    \noalign{\smallskip}\hline
    \end{tabular}
    \caption{\textit{Central Processing Unit (CPU)} and \textit{Graphics Processing Unit (GPU)} specifications for the utilized devices \textit{``PC"} and \textit{``Laptop"} in the evaluation process.}
    \label{tab:DeviceSpecification}
\end{table}

\subsubsection{Quantitative Results}\label{sec:quant_results}

Regarding the deployed YOLO models, the corresponding mAP values for the training (``train") and performance evaluation (``eval") are shown in Table~\ref{tab:YOLO_TrainEval}.
\begin{table}[!ht]
    \centering
    \begin{tabular}{lcccccc}
    \hline\noalign{\smallskip}
     YOLO model & -v3n & -v5n  & -v8n  & -v3x & -v5x & -v8x \\
     \noalign{\smallskip}\hline\noalign{\smallskip}
     train mAP.5     & 0.995 & 0.995 & 0.995 & 0.995 & 0.995 & 0.995 \\
     train mAP.5:.95 & 0.935 & 0.970 & 0.970 & 0.984 & 0.984 & 0.986 \\
     eval mAP.5      & 0.994 & 0.994 & 0.994 & 0.993 & 0.993 & 0.993 \\
     eval mAP.5:.95  & 0.945 & 0.974 & 0.975 & 0.983 & 0.983 & 0.985 \\
     \noalign{\smallskip}\hline
    \end{tabular}
    \caption{Mean Average Precision (mAP) metrics for different YOLO models and architecture sizes, automatically returned by the framework after the final training epoch (train mAP) and afterwards computed during the evaluation process (eval mAP).}
    \label{tab:YOLO_TrainEval}
\end{table}
The results indicate that both YOLOv5n and YOLOv8n perform on the same level, as do their respective -x counterparts. Only YOLOv3n shows a significantly lower mAP@[.5:.95]. Across all versions, the extra-large models yield higher mAP@[.5:.95] values than nano models, while the mAP@[.5] remains nearly identical. These results confirm the benefit of evaluating both mAP values, since mAP@[.5] only covers detection and mAP@[.5:.95] additionally captures localization accuracy, which is particularly important for varying swimmer postures and orientations. Further improvements might be achieved by dataset extension and extending the training process, as the method of \textit{early stopping} terminates the process when no improvement is made over the last epochs. However, it is a thin line to avoid \textit{overfitting}.\par%

To evaluate real-time applicability of the detection framework, we measured the \textit{Time Per Image (TPI)}, comprising the total processing time per image, which was automatically obtained from the framework. We derived the corresponding \textit{frame rate} in FPS as reciprocal from the TPI (rounded down), as displayed for the trained YOLO networks on the GPU and CPU devices ``PC" and ``Laptop" in Table~\ref{tab:YOLO_TPI_FPSn} and Table~\ref{tab:YOLO_TPI_FPSx} for the nano and extra-large architecture, respectively. In this context, real-time performance is associated with a frame rate of at least 30 FPS \citep{FPS1, FPS2, FPS3}.
\begin{table}[!ht]
    \centering
    \begin{tabular}{lrrrrrr}
    \hline\noalign{\smallskip}
    \multirow{2}{*}{Model} & \multicolumn{2}{c}{YOLOv3n} & \multicolumn{2}{c}{YOLOv5n} & \multicolumn{2}{c}{YOLOv8n}  \\
    & \multicolumn{1}{l}{TPI} & \multicolumn{1}{l}{FPS} & \multicolumn{1}{l}{TPI} & \multicolumn{1}{l}{FPS} & \multicolumn{1}{l}{TPI} & \multicolumn{1}{l}{FPS} \\
    \noalign{\smallskip}\hline\noalign{\smallskip}
     PC GPU     & 3.0ms & 333 & 1.8ms  & 555 & 1.8ms  & 555 \\
     PC CPU     & 16.6ms & 60 & 51.3ms & 19  & 56.0ms & 17  \\
     LAPTOP GPU & 9.7ms & 103 & 7.7ms  & 129 & 6.4ms  & 156 \\
     LAPTOP CPU & 79.5ms & 12 & 64.5ms & 15  & 80.1ms & 12  \\
     \noalign{\smallskip}\hline
    \end{tabular}
    \caption{Total processing TPI in milliseconds (ms) and frame rate in FPS for different YOLO models and their smallest (nano) architecture sizes.}
    \label{tab:YOLO_TPI_FPSn}
\end{table}
\begin{table}[!ht]
    \centering
    \begin{tabular}{lrrrrrr}
    \hline\noalign{\smallskip}
    \multirow{2}{*}{Model} & \multicolumn{2}{c}{YOLOv3x} & \multicolumn{2}{c}{YOLOv5x} & \multicolumn{2}{c}{YOLOv8x}  \\
    & \multicolumn{1}{l}{TPI} & \multicolumn{1}{l}{FPS} & \multicolumn{1}{l}{TPI} & \multicolumn{1}{l}{FPS} & \multicolumn{1}{l}{TPI} & \multicolumn{1}{l}{FPS} \\
     \noalign{\smallskip}\hline\noalign{\smallskip}
     PC GPU      & 14.3ms  & 69 & 19.0ms & 52 & 22.2ms & 45  \\
     PC CPU      & 100.7ms & 9  & 125.3ms & 7 & 1003.8ms & 0 \\
     LAPTOP GPU  & 57.1ms  & 17 & 74.2ms & 13 & 80.3ms & 12  \\
     LAPTOP CPU  & 580.1ms & 1  & 780.3ms & 1 & 1624.3ms & 0 \\
     \noalign{\smallskip}\hline
    \end{tabular}
    \caption{Total processing TPI in milliseconds (ms) and frame rate in FPS for different YOLO models and their largest (extra-large) architecture sizes.}
    \label{tab:YOLO_TPI_FPSx}
\end{table}
The results indicate a very strong performance for the nano architecture of all YOLO models on the GPU for both PC and Laptop with a frame rate of $103$ to $555$ FPS. On the PC, even the extra-large architecture achieves real-time performance on GPU, but their performance drops below real-time on the laptop. On CPU, no architecture except YOLOv3n reached real-time performance, revealing the importance of GPU employment.\par%

Overall, the extra-large models offer superior detection and localization accuracy but require considerably more computational resources. When a centralized, high-performance device is available, extra-large architectures are arguably the best option. However, this requires permanent data transfer between the centralized device and the UAV, which may introduce bandwidth and latency limitations. In comparison, nano architectures are less accurate but sufficiently lightweight to be run directly on a UAV. This makes them a practical alternative for scenarios involving UAV fleets or constrained communication, ensuring responsiveness and scalability through decentralized on-board processing. 

\subsubsection{Swimmer Localization and Alert Definition}\label{sec:swimmer-localization}

\begin{figure*}[!ht]
    \centering
    \includegraphics[width=\textwidth]{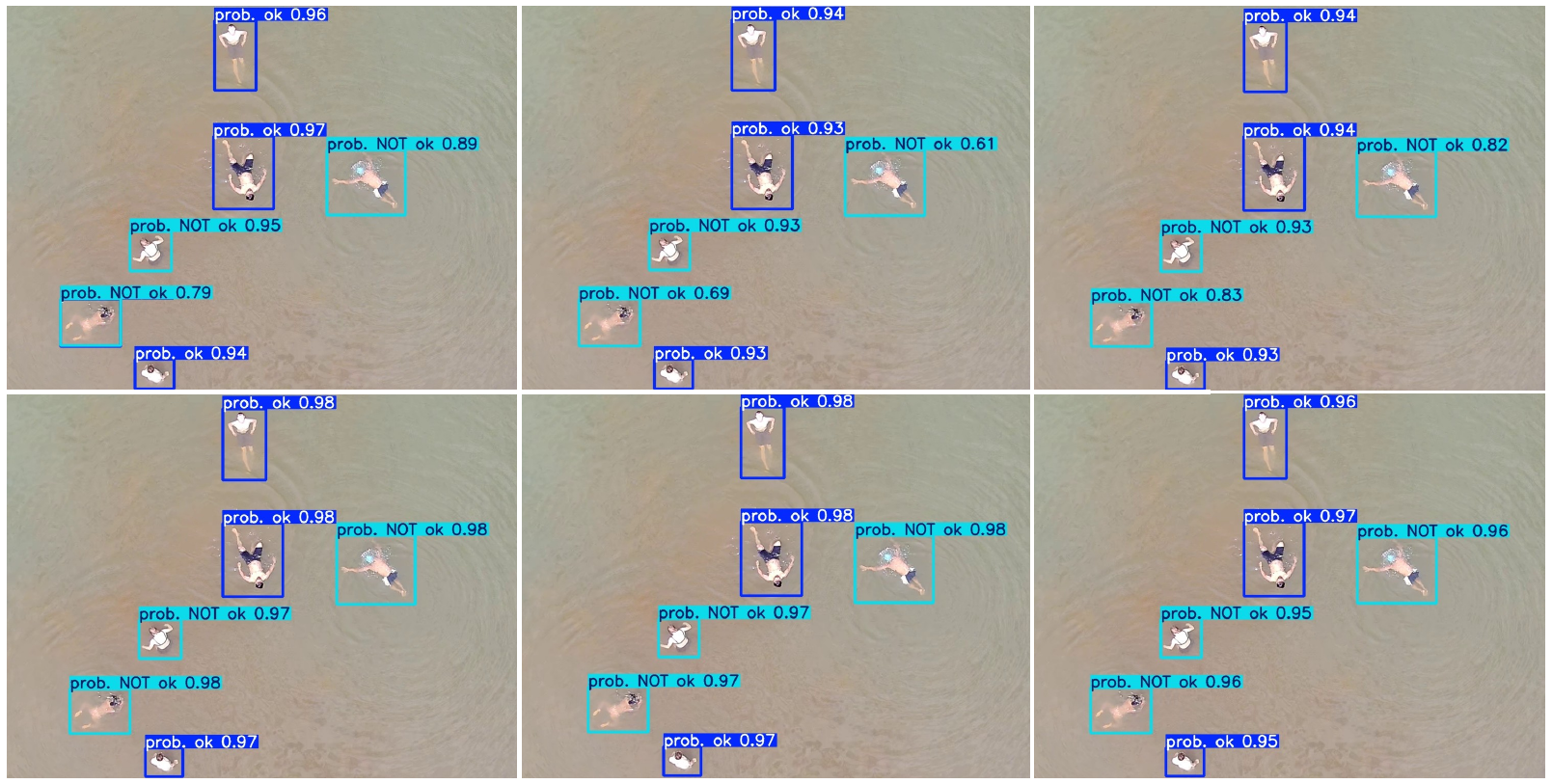}
    \caption{Detection results for all trained YOLO models and architectures. The top row shows nano variants (YOLOv3n, -v5n, -v8n), while the bottom row presents extra-large variants (YOLOv3x, -v5x, -v8x).}
    \label{fig:YOLOevalCollage}
\end{figure*}

After the quantitative evaluation of YOLO performance metrics, we now address a qualitative evaluation by showing and discussing representative detection results. Figure~\ref{fig:YOLOevalCollage} shows six detection results of the same scene using all trained YOLO versions and their respective architectures. The upper row corresponds to the nano architectures and the lower row to the extra-large architectures, respectively.
At first glance, the images indicate that the class ``prob. ok" is consistently detected with a high confidence, with a minimum of $0.93$ for nano models to $0.95$ for their extra-large counterparts. More relevant for our study is the detection of the ``prob. NOT ok" class, where the nano versions only reach a minimum confidence $0.61$, while the extra-large counterparts achieve $0.95$. This discrepancy might be explained by the limited number of training samples for the ``prob. NOT ok" class, reducing the robustness of the lightweight models. Thus, in terms of accuracy, Figure~\ref{fig:YOLOevalCollage} again emphasizes the superiority of extra-large architectures for this critical task. Interestingly, YOLOv3n outperforms YOLOv5n in this case, despite lower overall mAP values in the quantitative evaluation. However, this does not apply for all scenes.\par
\begin{figure*}[!ht]
    \centering
    \includegraphics[width=\textwidth]{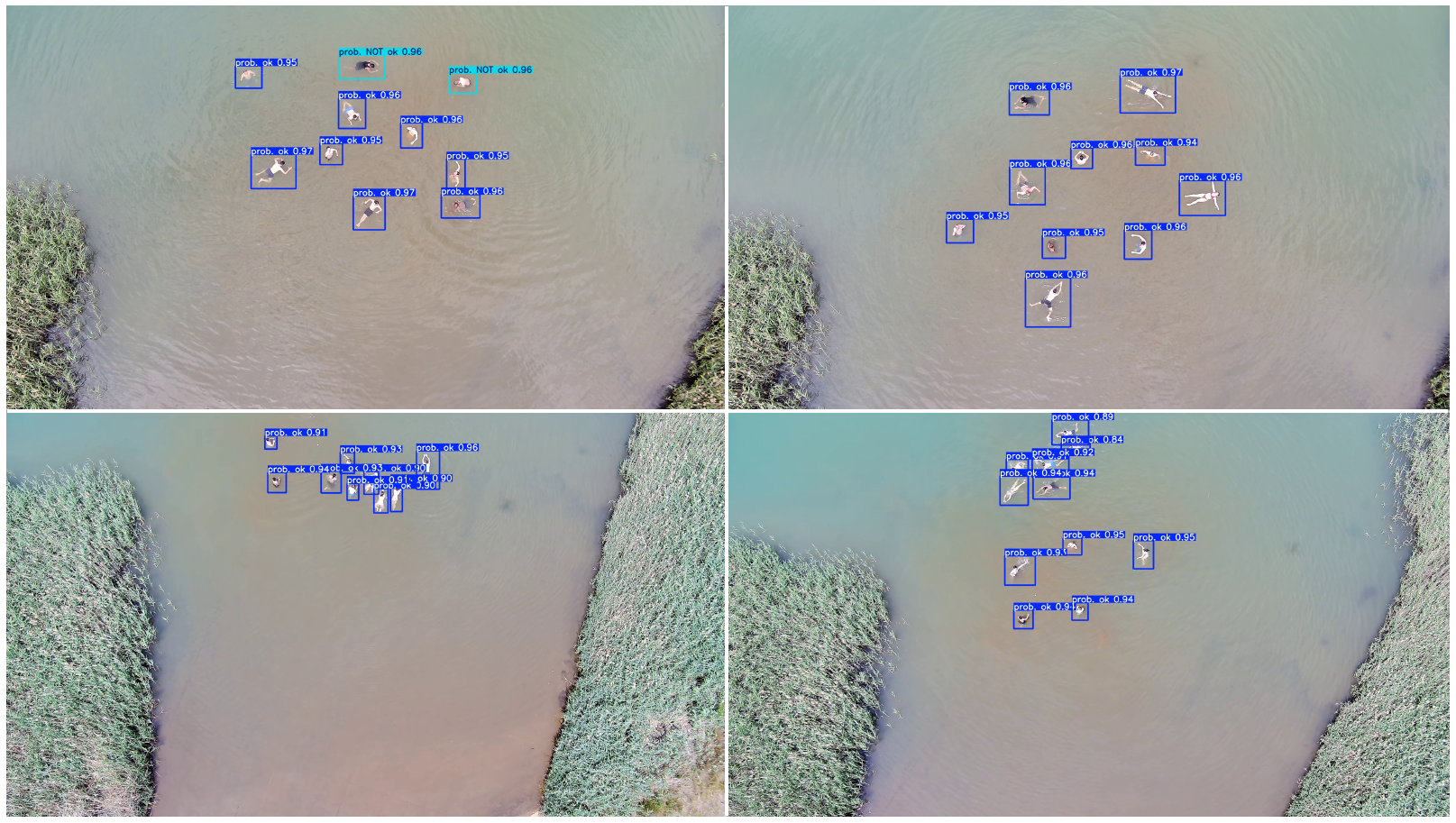}
    \caption{Detection results with YOLOv8x in a scene containing multiple swimmers, some in close proximity, being successfully distinguished and classified by the model.}
    \label{fig:YOLOevalManySwimmers}
\end{figure*}

\begin{figure}[!ht]
    \centering
    \subfloat[A false positive detection caused by light reflections at the bottom of the image.]{%
        \includegraphics[width=0.48\textwidth]{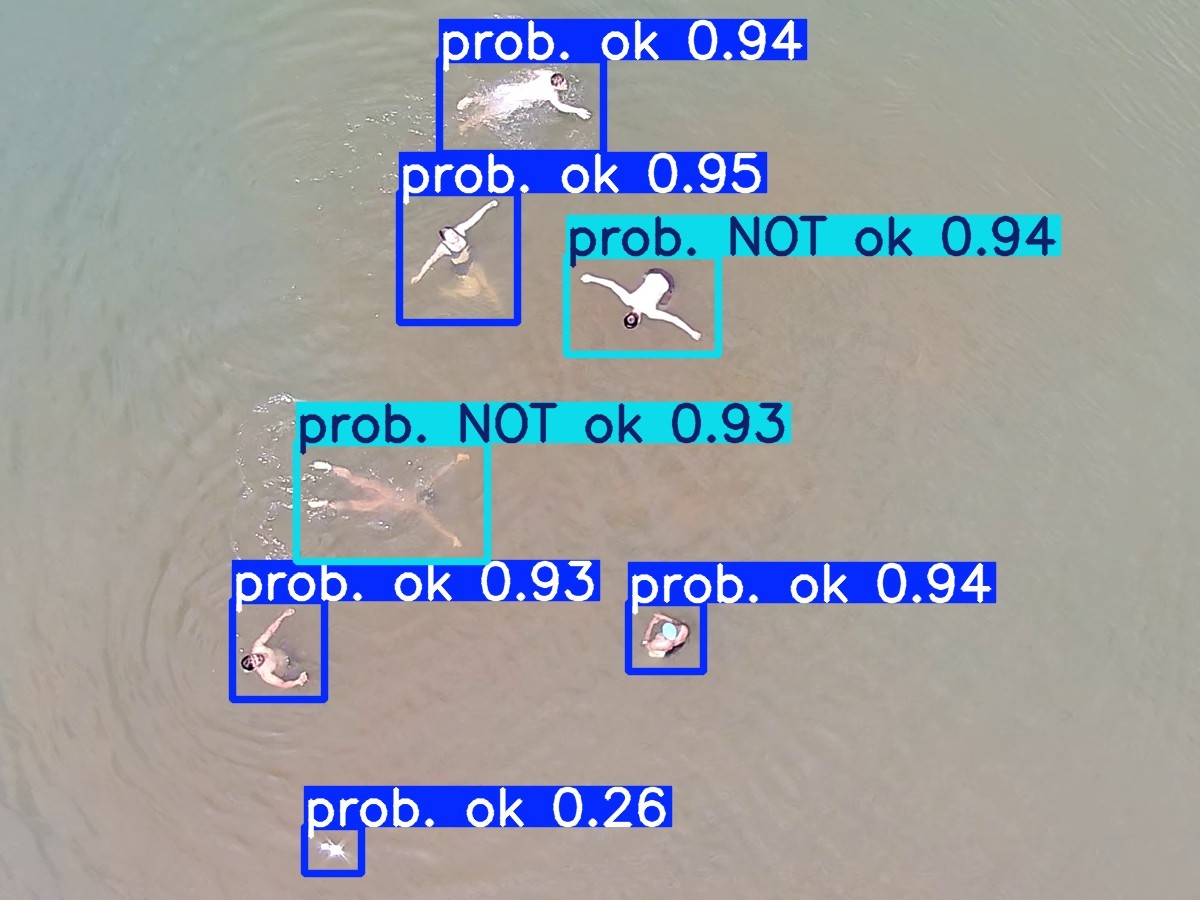}
        \label{fig:YOLOeval_v8n}
    }
    \hfill
    \subfloat[Unclear classification of a swimmer as ``prob. ok" and ``prob. NOT ok".]{%
        \includegraphics[width=0.48\textwidth]{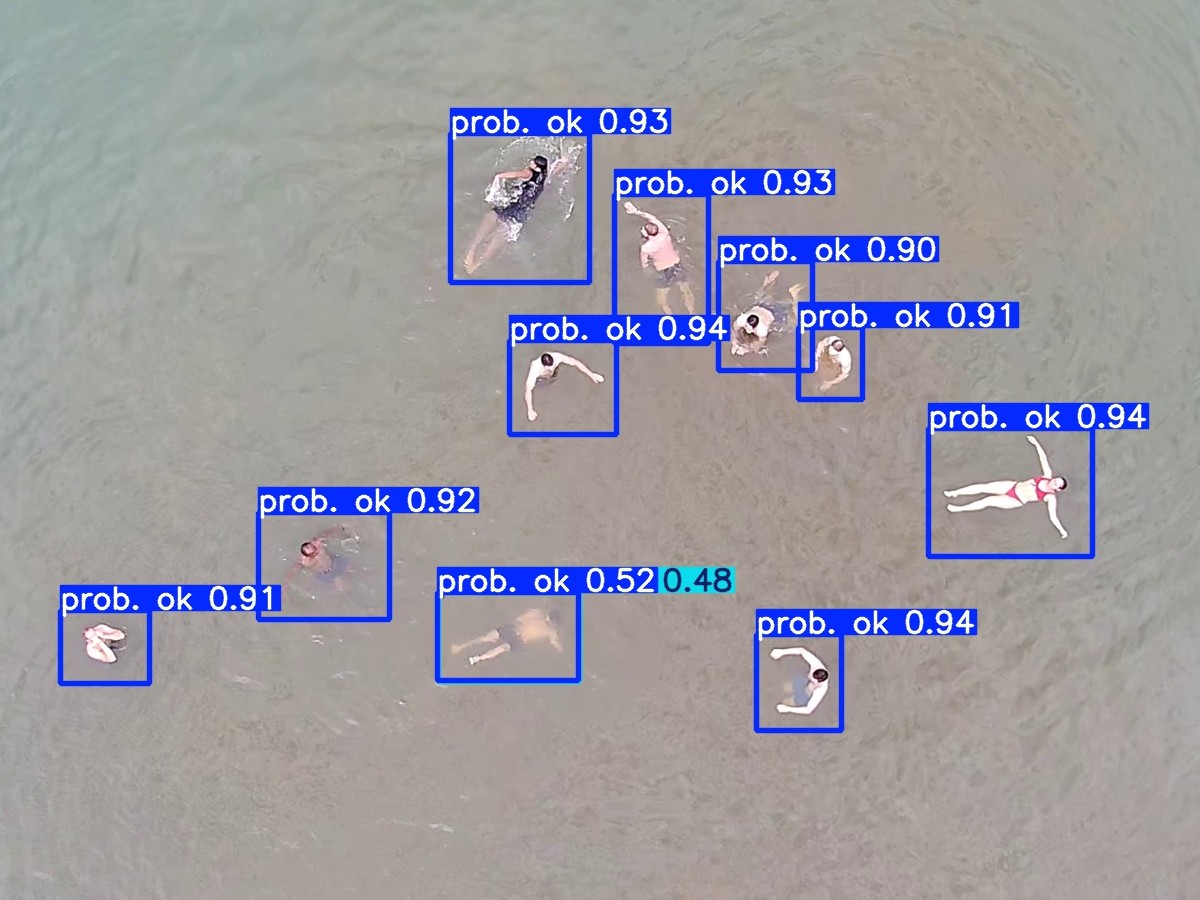}
        \label{fig:YOLOeval_v5n}
    }
    \caption{Examples of failure cases of the YOLOv5n model.}
    \label{fig:yolov5n_fails}
\end{figure}

%
Another common challenge in object detection is to distinguish objects in close proximity. The trained models can reliably detect multiple swimmers very close within a group, even when bounding boxes overlap. This is illustrated in Figure~\ref{fig:YOLOevalManySwimmers}, employing YOLOv8x. Importantly, these images show no false classifications or false localizations, as only swimmers are detected, while other surface structures such as water reflections and vegetation (reed) are correctly excluded. 
\begin{figure*}[!ht]
    \centering
    \includegraphics[width=\textwidth]{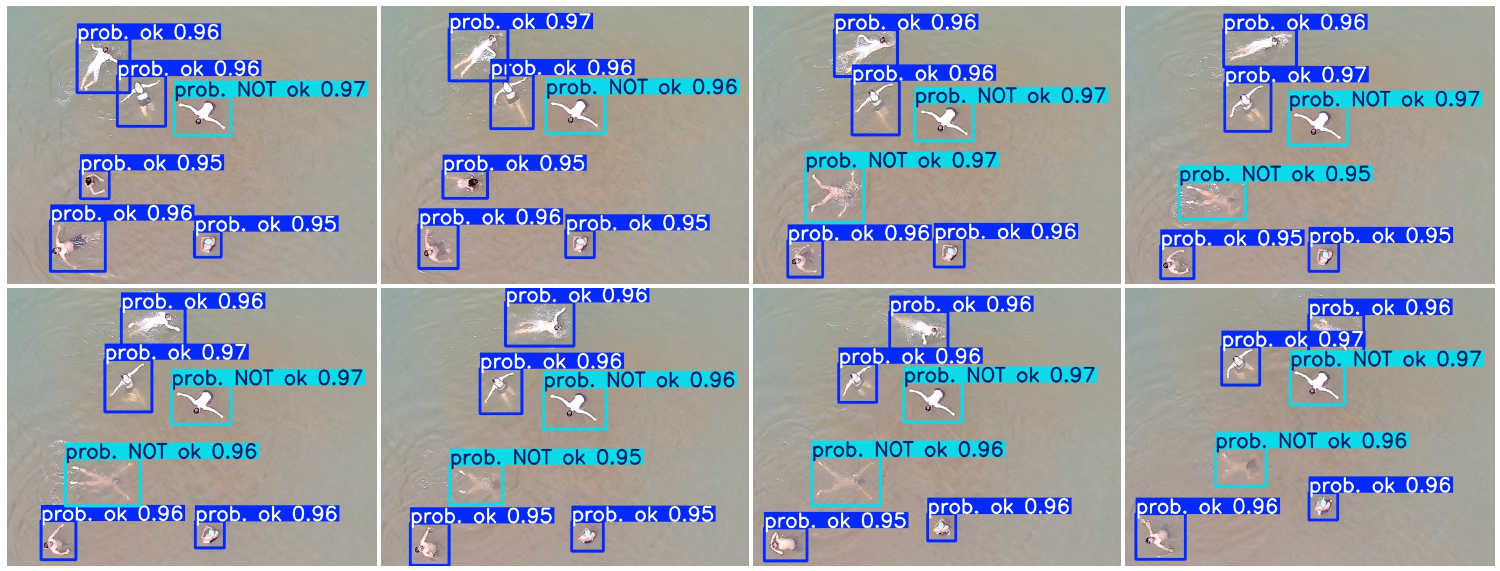}
    \caption{Sequence of detection results with YOLOv8n, illustrating the transition of a swimmer's classification from \textit{"prob. ok"} to \textit{"prob. NOT ok"}, highlighting the model’s ability to capture dynamic behavioral changes.}
    \label{fig:YOLOevalSequence}
\end{figure*}
A particularly important aspect of our work involves the detection of the \textit{"prob. NOT ok"} class and the ability to recognize transitions between the classes. 
Figure~\ref{fig:YOLOevalSequence} shows a sequence of frames (from top left to bottom right) where four people are swimming regularly, one person is floating without movement, which is correctly identified as ``prob. NOT ok", and a swimmer is classified as ``prob. ok" in the first two frames, then begins to dive, after which the label correctly switches to ``prob. NOT ok" for the remainder of the sequence. Based on the training data, this is a very solid result, demonstrating the potential of the models to capture dynamic changes in swimmer behavior. Building on that, a robust alerting strategy could be implemented by evaluating recent classifications in a sliding window (e.g. 30 frames). If a ratio of detections is classified as ``prob. NOT ok", the swimmer could be flagged as potentially in distress, even if frames occasionally produce ``prob. ok" labels. Nonetheless, some limitations are evident. Figure~\ref{fig:YOLOeval_v8n} shows a false positive, caused by light reflections on the water surface, mistakenly captured by YOLOv5n. Additionally, Figure~\ref{fig:YOLOeval_v5n} shows an unclear detection, classifying a swimmer as both ``prob. ok" and ``prob. NOT ok" simultaneously. This phenomenons might be related to the nano architecture, and increasing the detection threshold  solves these issues. Furthermore, in extreme cases of light reflection, employing further pre-processing steps can reduce the impact and enhance detection robustness.

\subsection{Simulation}\label{sec:sim_seenplatte}

In the following, we apply the simulation frameworks from Sections \ref{sec:sim-model_SRO} and \ref{sec:sim-model} to the test area in the \textit{Lusatian Lake District} which is specified by the black outline in Figure~\ref{fig:demandexpert} and is referred to as operational area.

\subsubsection{Standard Rescue Operation Simulation}\label{sec:SRO_sim_seenplatte}

\begin{figure}[ht]
    \centering
    \includegraphics[width=1.0\textwidth]{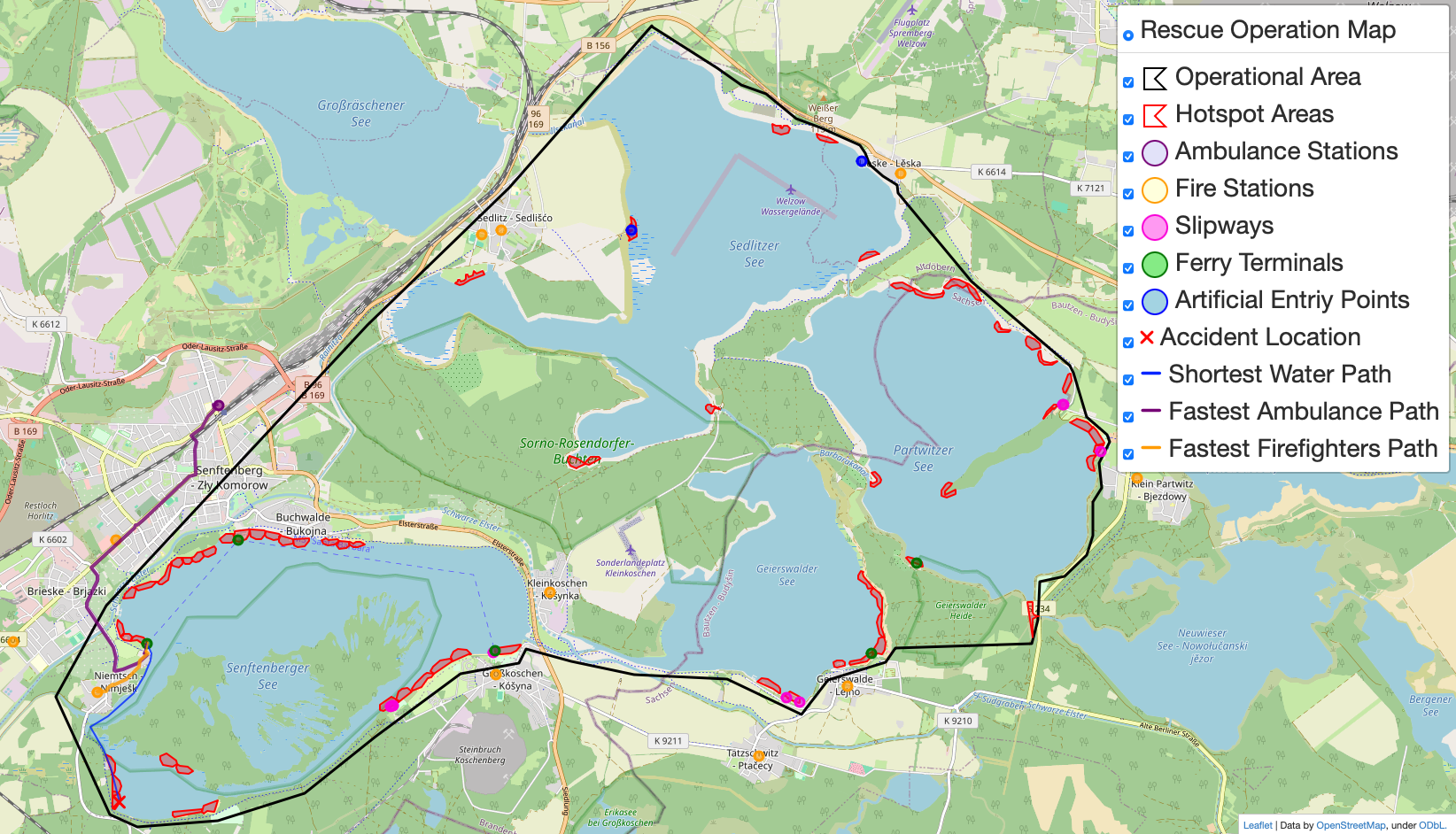}
    \caption{Illustration of an exemplary water rescue SRO in the Lusatian Lake District, Germany including the fastest path of the nearest ambulance (purple line), fire truck (orange line) and life boat (blue line) as well as predefined hot spot areas (red areas).}\label{fig:demandexpert}
\end{figure}

First, we need to define the sets of fire stations $\mathcal{F}$, rescue stations $\mathcal{R}$ and water access points $\mathcal{W}$. We load the rescue stations and fire stations from OSM (key-tag pairs ``amenity = fire\_station" and ``emergency = ambulance\_station") within a radius of 20 and 2 kilometers, respectively, from the operational area. Furthermore, we load potential water access points from OSM, using slipways and ferry terminals (key-tag pairs ``leisure = slipway" and ``amenity = ferry\_terminal"). Since no such points are marked on OSM for one of the lakes, Lake Sedlitz, we added artificial water access points there in consultation with local authorities. All determined water access points are shown in Figure~\ref{fig:demandexpert} in prink, green, or blue. To determine travel distances $\delta_{r,w}$ and $\tilde{\delta}_{f,w}$ we compute the fastest paths from emergency stations $r\in\mathcal{R}$ and fire stations $f\in\mathcal{F}$ to the water access points $w\in\mathcal{W}$ by loading the street network from OSM, finding the respective nearest nodes on the graph and running \textit{Dijkstra's algorithm} \cite{Dijkstra1959} using the associated travel times as edge weights that are determined by the length and speed limit of the associated road section, which may be exceeded by ambulances and fire trucks. According to the findings of \citet{pappinen2022driving}, the respective speed limits may be exceeded for urgent emergency services, but by no more than 30\% on average. As this is a very urgent operation, we assume a 30\% increased speed limit to calculate the travel times for every road. Since the nearest node of the OSM street network may be some distance away from the water access point, we determine the Euclidean distance, set a speed of 10 km/h for this additional distance to be covered on foot and add it to the travel time to ultimately obtain $\delta_{r,w}$ and $\tilde{\delta}_{f,w}$. We calculate travel times $\bar{\delta}_{w, \bm{\tau}}$ from the nearest water access point $w\in\mathcal{W}$ to the accident site $\bm{\tau}$ by calculating the shortest path using the \textit{visibility graph algorithm} developed by \citet{lee1978proximity} to guarantee staying on the water surface and we use \textit{Dijkstra's algorithm} \cite{Dijkstra1959} to find the shortest path. Firefighting and rescue boats in Germany must comply with German standard DIN 14961 \cite{DIN14961_2013}, defining its technical requirements. Of the various boat models, we selected the highest maximum speed we could find, 70 kilometers per hour for an upgraded high-tech boat \cite{mehrzweckboot}, as a reference speed to finally compute boat travel times $\bar{\delta}_{w, \bm{\tau}}$. Note that this is a very optimistic estimate, as most lifeboats with equipment and several people on board do not reach this maximum speed in reality.

We determine the preparation times $\vartheta_F, \vartheta_R$ of the firefighters and parame\-dics until departure from their respective stations. Since paramedics are located directly at the rescue station, we assume $\vartheta_R=0$. The fire departments in the region are exclusively volunteer fire departments, which is why firefighters are usually not directly at the fire station but available on demand via radio. After being alerted, they rush to the fire department. Unfortunately, there is no good quality data available on the preparation times for volunteer fire departments, as data availability in EMS is generally a problem. There are also regional differences, as each federal state has its own laws for emergency services and firefighting. We assume a truncated normal distribution $\vartheta_R \sim \mathcal{TN}(\mu, \sigma^2; a, b)$ with mean $\mu = 120$, variance $\sigma^2 = 30$ and limits $a=0$ and $b=240$ (in seconds). This is on average slightly less than local experts estimate, as all conventions for determining the variables are rather optimistic, which means that the simulation will make an optimistic estimate on the target detection time.\par%

\begin{figure}[ht]
    \centering
    \includegraphics[width=0.8\textwidth]{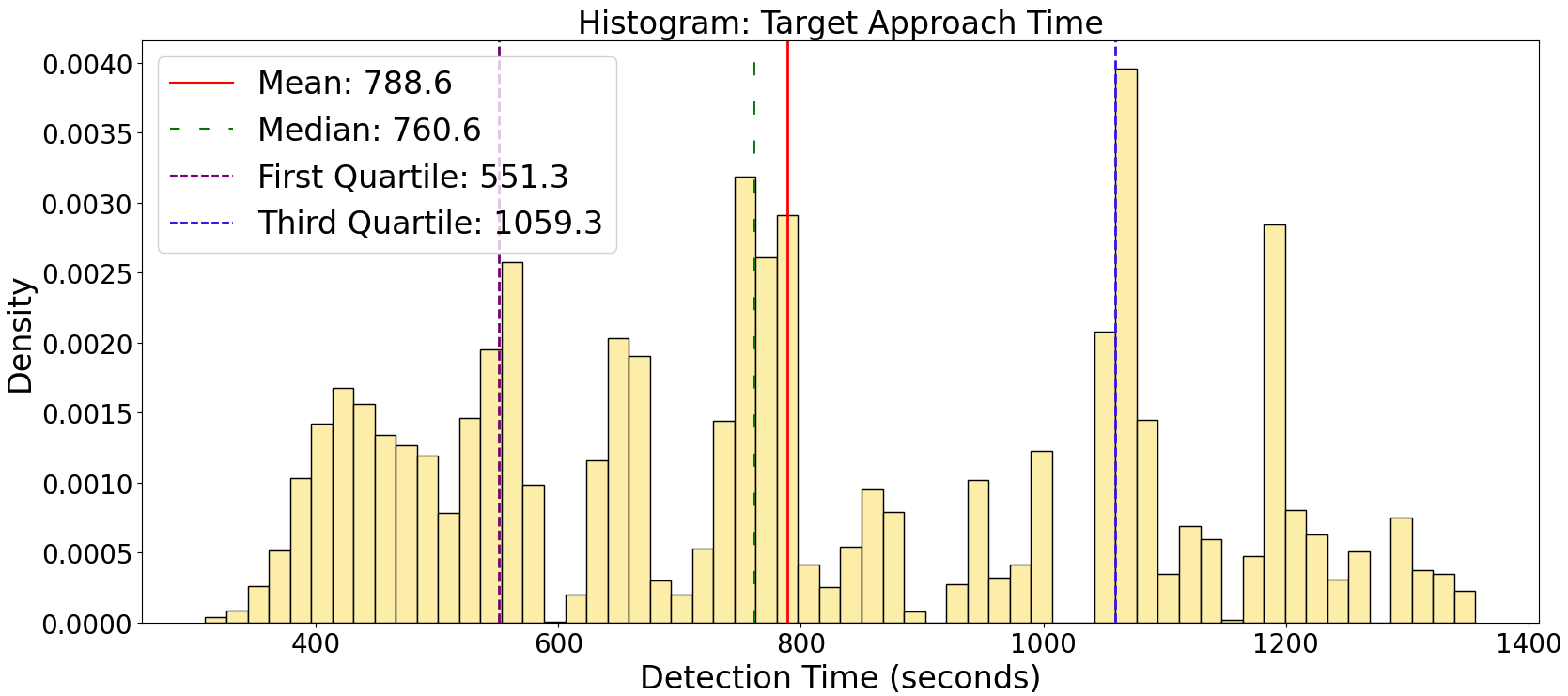}
    \caption{Histogram of the estimated target approach time for a SRO of a distressed swimmer, based on a MCS with $N^{\text{MCS}}=10,000$ iterations.}\label{fig:monte_carlo_SRO}
\end{figure}

An example SRO is depicted in Figure \ref{fig:demandexpert}, including the fastest paths of the nearest ambulance (violet line) and the nearest fire brigade (orange line) and the shortest path on water from water access point to swimmer location (blue line). Finally, we perform a MCS with $N^{\text{MCS}}=10,000$ runs. To this end, we consulted experts to define several hotspot areas where swimming is particularly popular. We then used OSM to define weights and assign them to each hotspot area, as detailed by \citet{Zell2024}. Those hotspot areas are depicted in Fig~\ref{fig:demandexpert}.
Now, in each of the $N^{\text{MCS}}$ runs, a hotspot area is sampled randomly according to the weighting of the areas in which the target is then randomly placed following a uniform distribution. The results of the MCS are summarized with a histogram of the target approach time in Fig~\ref{fig:monte_carlo_SRO}. It can be seen that the target approach time does not follow a unimodal distribution and varies widely. It always takes at least 150 seconds, with a mean of roundabout 13 minutes, but can also increase to almost 23 minutes, depending heavily on the specific location of the emergency.

\subsubsection{UAV-based Water Rescue Simulation}\label{sec:UAV_sim_seenplatte}

The simulation approach presented in Section \ref{sec:sim} is now use to estimate speed and reliability of locating distressed swimmers with a specific UAS configuration and search algorithm. For this, we use performance indicators average finite detection time $\bar{t}_m$ target detection rate $\chi_m\in[0,1]$ from Equations \eqref{eq:expecation_average} and  \eqref{eq:success_ratio}, respectively. These performance indicators refer to the response time and the success rate.\par%

In the simulation, we consider the predefined target areas from Figure \ref{fig:demandexpert}:  popular swimming areas, predefined in consultation with local experts. Within these areas, the distressed swimmer is assumed to be stationary over time and is placed at random (following a uniform distribution over the search area). We can then determine whether and how quickly the target is found by the UAS. This experiment is repeated $N^{\text{MCS}}$ times to enable a Monte Carlo simulation (MCS).\par%
We employ the location-allocation optimization MILP from Part 1 of this paper series for UAS configuration planning, determining the hangar locations and UAV allocations for the simulation. Similarly, we employ the in-advance offline S\&R flight planning MILP from Part 1 of this series as a search algorithm. The aforementioned models are also described by \citet{Zell2024}. Since some of the search areas are up to several minutes away from the UAS locations,  we split the S\&R operation into two parts: The approach flight from the hangar to the target area and the search flight within the target area. First, we determine the point on the edge of the target area that is closest to the hangar and approach it first. Next, this point serves as the starting point for the search flight. This implies that, instead of solving one large MILP for the entire mission in a designated search area, a very small MILP must be solved for the approach flight, not containing any waypoints, and a moderately large MILP must be solved for the search flight. This can still be computationally intensive, but not nearly as much as without splitting into two MILPs. We performed the simulation for various instances, namely for the solutions of the 
\textit{Capacitated Cooperative Maximum Covering Location Problem (CCMCLP)} with   $p=1,\ldots, 6$ homogeneous hangars to be set up containing a single UAV from a homogeneous fleet (cp. first instance of \citet{Zell2024}). In each experiment, $N^{\text{MCS}} = 100000$ S\&R operations were performed. Note that only those search areas within the range of the available battery power of the UAS are considered, as determined by the CCMCLP. Other areas are excluded from the search and associated targets are declared untraceable.\par%

The results of the simulations are logical: the more resources deployed, the better the detection rate and average detection time. Table \ref{tab:sim_results} summarizes the results for different UAS configurations as indicated by the $p$ value for numbers of hangars used. From that, it seems that two UAVs can cover the area almost completely with a success rate of more than 99 \%, but there is still a lot to gain in terms of average detection time.
\begin{table}[!ht]
    \centering
    \begin{tabular}{l|l|l|l}
     p & Average detection time $\bar{t}_{p} [s]$  & Success rate $\chi_{p} [\%]$ & Simulation run time $[s]$\\[0.5ex]
     %  &  & \rule{0pt}{0.1ex}  &  &  &  \\
    \hline
    1 & $371.82$ & $57.31$ & $28.79$ \\
    2 & $155.88$ & $99.39$ &  $76.07$ \\
    3 & $123.94$ & $99.64$ & $103.64$ \\
    4 & $100.47$ & $99.53$ & $144.15$ \\
    5 & $101.56$ & $99.50$ & $174.91$ \\
    6 & $94.09$ & $99.54$ & $204.15$ \\
    \end{tabular}
    \caption{UAS-based water rescue simulation results including the average detection time (in seconds), detection rate (in percent) and computing times such as the time to solve the location-allocation MILP, the average time to solve the Search \& Rescue MILP and the computing time of the MCS with $100,000$ runs on a MacBook Pro (Intel Core i9 running 1 thread at 2.4 GHz clock speed and 64 GB RAM).}
    \label{tab:sim_results}
\end{table}
%

\begin{figure}[ht]
    \centering
    \subfloat[$p=1$]{%
        \includegraphics[width=0.49\textwidth]{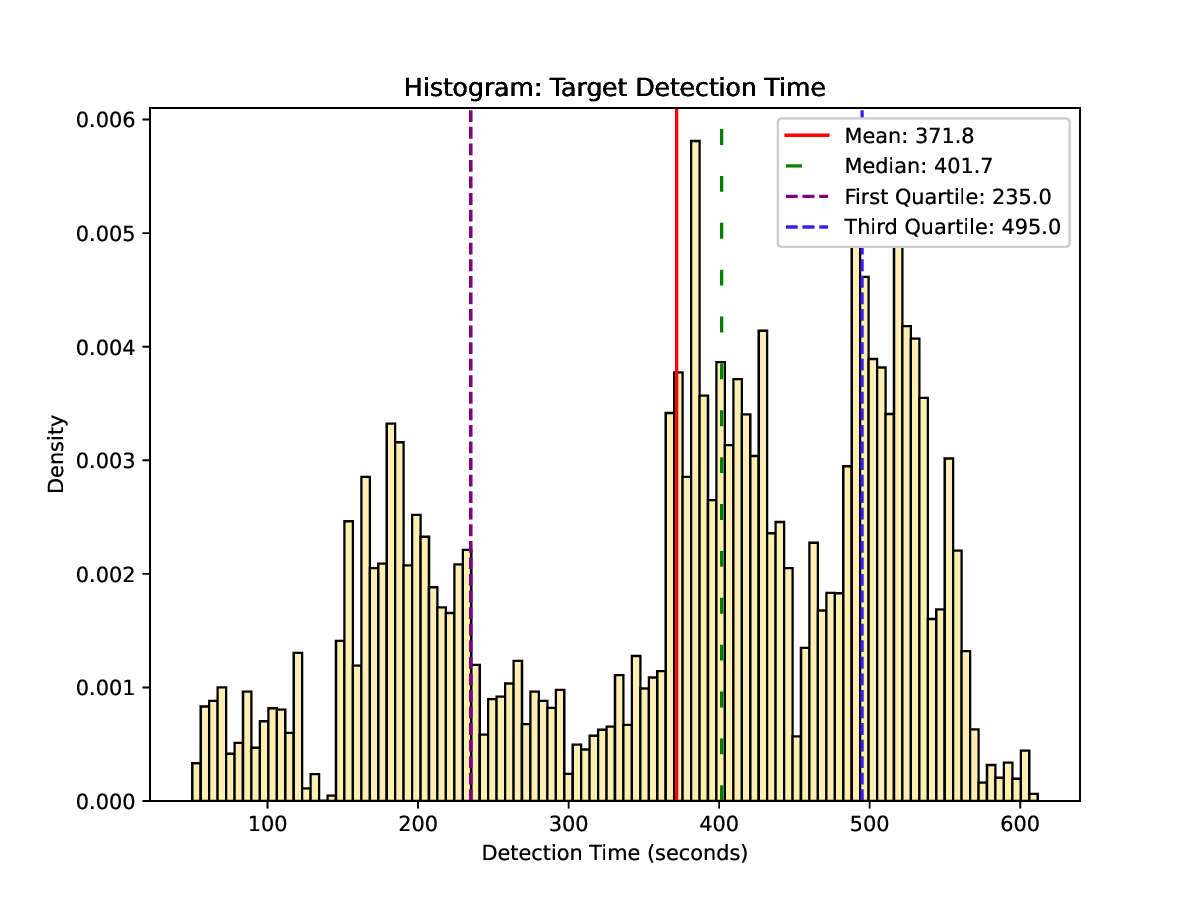}
        \label{fig:det_time_p1}
    }
    \hfill
    \subfloat[$p=2$]{%
        \includegraphics[width=0.49\textwidth]{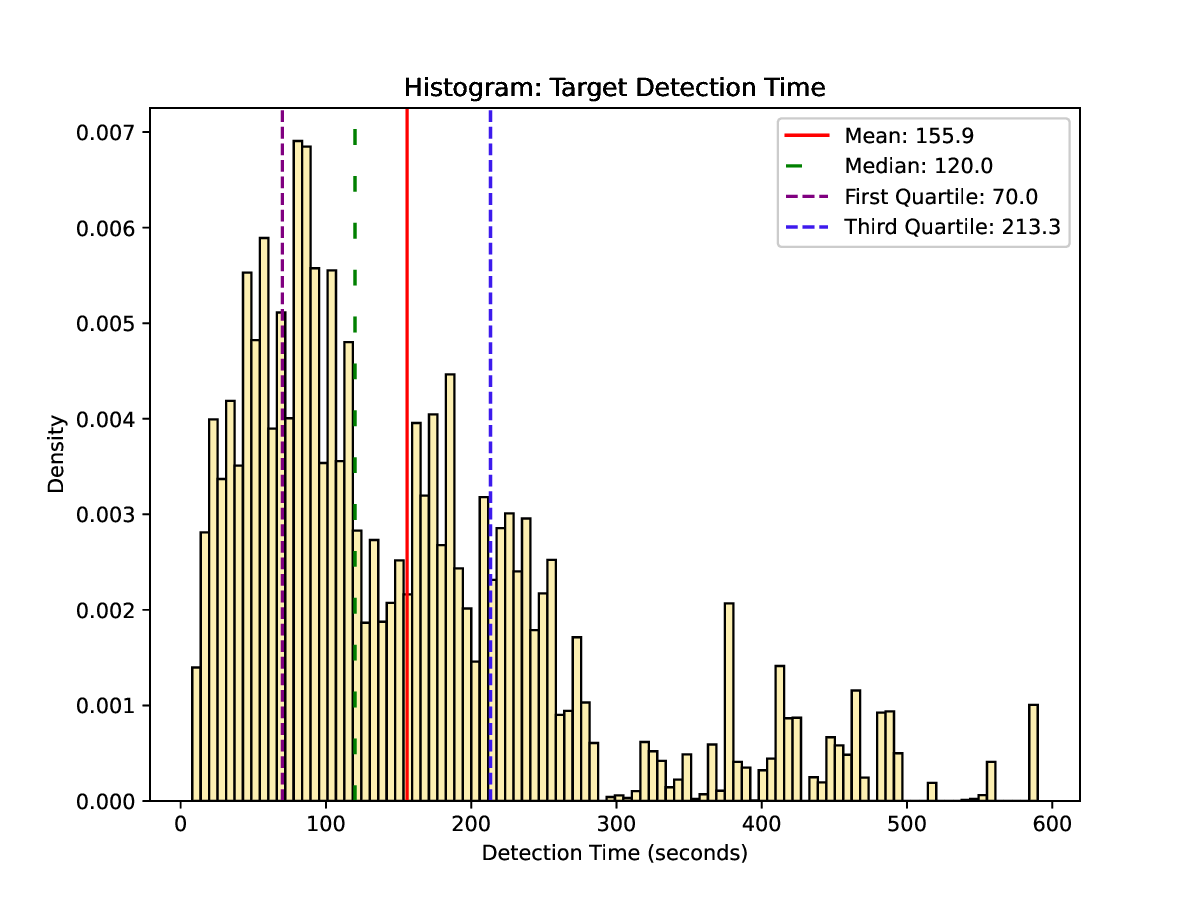}
        \label{fig:det_time_p2}
    }

    \subfloat[$p=3$]{%
        \includegraphics[width=0.48\textwidth]{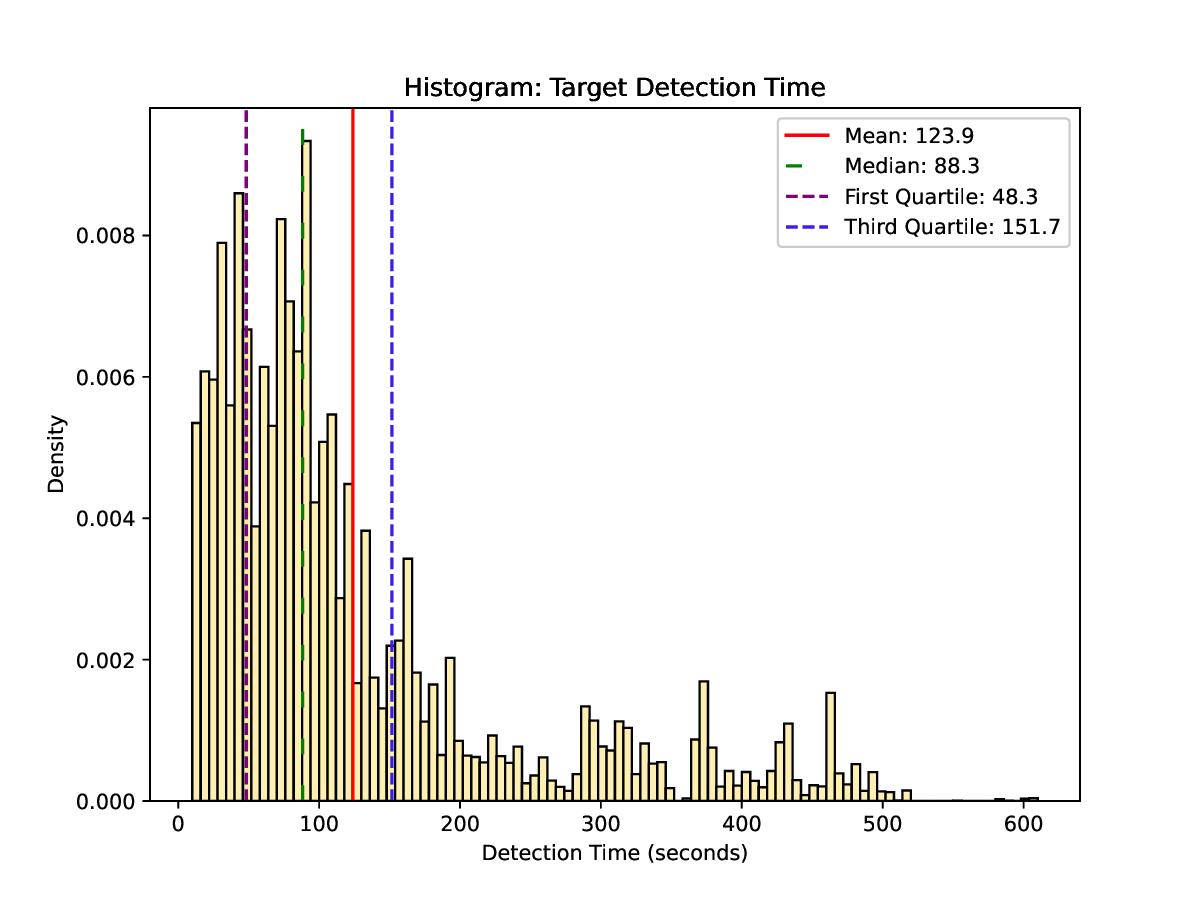}
        \label{fig:det_time_p3}
    }\hfill
    \subfloat[$p = 6$]{%
        \includegraphics[width=0.48\textwidth]{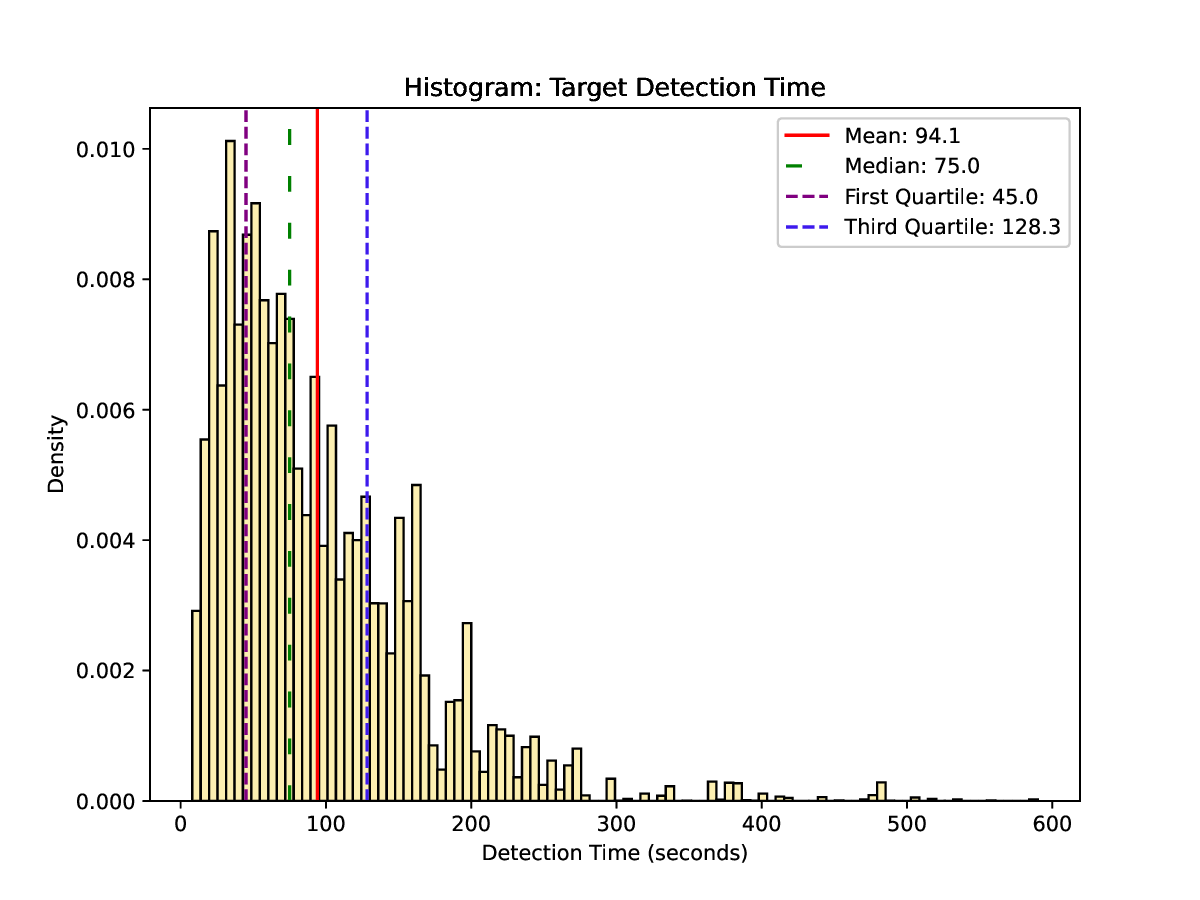}
        \label{fig:det_time_p6}
    }
    \caption{Histograms for the simulation of 100,000 S\&R operations using UAS configurations with different numbers of hangars $p$ and respective success rates of $\chi_{(p=1)} = 57.3\%$, $\chi_{(p=2)} = 99.39\% $, $\chi_{(p=3)} = 99.64\% $, and $\chi_{(p=6)} = 99.54\% $.}
    \label{fig:sim_time_histograms}
\end{figure}

% simulation time histograms (p=1, p=3, p=6)
Figure \ref{fig:sim_time_histograms} depicts histograms of the simulated response or detection times for $p=1$, $p=2$, $p=3$ and $p=6$, respectively. The differences are remarkable: with only one hangar, it takes an average of about 6 minutes with a success rate of $57\%$, while with 6 hangars, around 90\% of the simulated missions are completed in less than 2 minutes, with a success rate of $99.54 \%$. Now, comparing the UAS-based results to the conventional SRO, with an average response time of around 13 minutes (cp. Figure~\ref{fig:monte_carlo_SRO}), the benefits of the UAS become clear.
%
% S&R MILP MIP Gaps & Run times
Although the computational effort required for the simulation is relatively modest, the overall computational effort per UAS configuration is very high, as a lot of computing time is required for location-allocation optimization and S\&R flight route optimization by solving the large associated MILPs. The computing time for the latter MILP is particularly significant, varying per scenario from an average of approximately 50 minutes to 74 minutes. For all target areas, the 65 hotspot areas from Figure \ref{fig:demandexpert} in our case, this adds up to 54 to 80 hours per scenario. This illustrates how the computational effort quickly accumulates for the entire framework, which is why we have limited our consideration in this paper to the six scenarios from Table \ref{tab:sim_results}. To maintain control over this, the instance sizes must be kept moderate, which involves careful adjustment of the model parameters. 

% ---------------------------------------------------------------
% Section: Conclusions
% ---------------------------------------------------------------

\section{Conclusions}\label{sec:conclusions}

In this paper, we presented a comprehensive framework for UAV-assisted water rescue in inland waters, based on an \textit{Unmanned Aerial System (UAS)} located on land near swimming areas, consisting of UAVs stored in purpose-built UAV hangars. Developed as a complement to conventional water rescue, the UAS can be deployed automatically to assist in the \textit{Search and Rescue (S\&R)} of distressed swimmers by scanning a search area with a visual camera using automatic swimmer recognition assistance and potentially dropping off a flotation device.\par%
Our work focused on two main aspects: a toolkit for automatic real-time image detection of distressed swimmers and a simulation-based assessment of water rescue operations. Both were applied within a case study in the \textit{Lusatian Lake District} in Eastern Germany as part of the research project \textit{RescueFly}.\par%

For the detection task, we addressed the challenge of creating a fully automated real-time identification system for distressed swimmers adopting the established \textit{You Only Look Once (YOLO)} method. Due to the scarcity of appropriate training data, we introduced a new UAV captured dataset allowing classification of swimmers as either ``probably okay" or ``probably not okay". The trained models showed solid real-time performance, measured by TPI and FPS, and high accuracy, evaluated through mAP.5 and mAP.5:.95, with architecture sizes tailored for lower-performance hardware. Nevertheless, further dataset expansion is necessary to achieve full automation, as the current dataset contains only simulated emergency behavior. We recommend extending coverage to location-specific ethnic groups and vulnerable age groups (children, seniors), different inland water environments and incorporating image sequence analysis, as reliable classification cannot be based exclusively on single frames. Ensuring robustness and reliability remains critical for such an important safety application.\par%

The simulation models developed in this study enable a direct comparison between Standard Rescue Operations (SROs) and UAS-assisted rescue of swimmers in inland waters in terms of rapid accessibility, which is the critical factor in water rescue. Using a Discrete Event Simulation (DES) approach, we modeled SROs following the standard procedure for inland water rescue for the \textit{Lusatian Lake District}, deploying the nearest ambulance and the nearest fire brigade to provide a life boat to approach the target. Further, we developed a generic DES-based simulation model compatible with any time-discrete UAV-based S\&R algorithm. By employing \textit{Monte Carlo Simulation (MCS)}, we estimated the mean target approach time, which is defined as the time until the first rescue resources (lifeboat or UAV) reaches the target and can provide a flotation device. We found that some highly frequented swimming areas are difficult to access quickly using SRO, resulting in target approach times of between 2.5 and 23 under optimistic assumptions regarding preparation and travel time. In contrast, even a UAS consisting of only two UAVs and two hangars achieved a substantial improvement in the test scenario for the mean target approach time by a factor of $5$ from roundabout $13$ minutes to about $2.5$ minutes. This demonstrates that a carefully and individually configured UAS is a valuable addition to the existing response system. Here, the performance of the UAS depends not only on the UAV and hangar specifications, but also significantly on demand and area size. For full-area coverage, the simulation framework provides a practical tool to estimate resource requirements, also benefiting from models developed in Part 1 of this paper series for in-advance flight route planning and location-allocation optimization. Possible extensions of the framework include incorporation of human behavior such as bystander intervention or movement of distressed swimmers and also improving the currently limited data basis for water rescue operations.

% closing paragraph 
Taken together, the automated real-time detection framework and sim\-u\-la\-tion-based evaluation framework form a valuable tool kit for enhancing swimmer safety in unsupervised inland water swimming areas. Real-time detection enables rapid identification of distressed swimmers, while simulation facilitates data-driven planning and optimized individual resource allocation, offering an integrated approach ranging from strategic resource deployment to real-time operational decision-making. The integration of the capabilities enables UAV-assisted water rescue operations to overcome access restrictions, shorten response times and therapy-free intervals, and ultimately improve survival chances in critical situations. Even though further refinements of models and datasets are needed, the presented framework contributes to a scalable and transferable basis towards future design, deployment and standardization of UAV-assisted water rescue in inland waters and beyond. Finally, we want to emphasize the importance of advancing this research area and developing solutions that make noticeable improvements in emergency situations. We strongly believe that collaboration between experts from different disciplines, combined with progressive testing and improvements, will ultimately lead to UAV-assisted water rescue making a significant impact on improving swimmer safety.
\section*{Acknowledgements}
The authors express their gratitude to IRLS Lausitz (Cottbus) and IRLS Ostsachsen (Hoyerswerda) for their productive collaboration in historical data collection as well as the whole \textit{RescueFly} project group for fruitful collaboration and inspiration.
\section*{Conflict of interest}
The authors have no competing interests to declare that are relevant to the content of this article.

\section*{Funding}
This study was funded by the German Federal Ministry for Digital and Transport (Bundesministerium für Digitales und Verkehr (BMDV)) within the project \emph{RescueFly} (reference number 45ILM1016C) as well as by the German Federal Ministry of Research, Technology and Space (Bundesministerium für
Forschung, Technologie und Raumfahrt (BMFTR)) within the project \emph{KI@MINT} (reference number 16DHBKI052).

\section*{Ethics approval}
The study was approved by the Ethics Committee of the Brandenburg University of Technology Cottbus-Senftenberg (protocol code: EK2024-07, date of approval: April 22, 2024). Informed consent was obtained from each individual for the part of the research involving human participants.

\section*{Data availability}
The authors declare, that upon reasonable request, the data and the code are available from the corresponding author.

\section*{Author contributions}

Conceptualization: Sascha Zell, Toni Schneidereit, Armin Fügenschuh, Michael Breuß;
Methodology: Sascha Zell, Toni Schneidereit, Armin Fügenschuh, Mi\-cha\-el Breuß;
Software: Sascha Zell, Toni Schneidereit;
Validation: Sascha Zell, Toni Schneidereit;
Formal analysis: Sascha Zell, Toni Schneidereit;
Investigation: Sascha Zell, Toni Schneidereit;
Resources: Sascha Zell;
Data curation: Sascha Zell, Toni Schneidereit;
Writing-original draft preparation: Sascha Zell, Toni Schneidereit;
Writing-review and editing: Armin Fügenschuh;
Visualization: Sascha Zell, Toni Schneidereit;
Supervision: Armin Fügenschuh, Michael Breuß;
Project administration:  Armin Fügenschuh, Michael Breuß;
Funding acquisition: Armin Fügenschuh, Michael Breuß

\bibliographystyle{unsrtnat}
\bibliography{literature}

% \newpage\null\thispagestyle{empty}\newpage
% \selectlanguage{english}
% \printbibliography

\end{document}